\documentclass[10pt,twocolumn,letterpaper]{article}

\usepackage[pagenumbers]{cvpr} 

\usepackage{graphicx}
\usepackage{amsmath}
\usepackage{amssymb}
\usepackage{booktabs}
\usepackage{multirow}
\usepackage{makecell}
\usepackage{bbm}
\usepackage{float}
\usepackage{xcolor}
\usepackage{algorithm}
\usepackage{algorithmic}
\usepackage{pifont} 
\usepackage{color, colortbl}
\usepackage{arydshln}
\usepackage{tabularx}
\usepackage{footnote}
\usepackage{nicefrac}
\usepackage{url}
\usepackage{bbm}
\definecolor{colorful}{rgb}{0.75, 0.93, 0.75}
\newcommand{\cmark}{\ding{51}}%
\newcommand{\xmark}{\ding{55}}%

\definecolor{cvprblue}{rgb}{0.21,0.49,0.74}
\usepackage[pagebackref,breaklinks,colorlinks,allcolors=cvprblue]{hyperref}

\def\paperID{} 
\def\confName{CVPR}
\def\confYear{2026}

\title{Recursive Think-Answer Process for LLMs and VLMs}

\author{Byung-Kwan Lee\thanks{Equal contribution} \hspace{0.02em} \thanks{Currently Research Scientist at NVIDIA}\\
KAIST\\
{\tt\small leebk@kaist.ac.kr}
\and
Youngchae Chee\footnotemark[1]\\
KAIST\\
{\tt\small litcoderr@kaist.ac.kr}
\and
Yong Man Ro\\
KAIST\\
{\tt\small ymro@kaist.ac.kr}
}

\begin{document}
\input{figures/figure_llm_bench}
\maketitle
\begin{abstract}
Think–Answer reasoners such as DeepSeek-R1 have made notable progress by leveraging interpretable internal reasoning. However, despite the frequent presence of self-reflective cues like ``Oops!'', they remain vulnerable to output errors during single-pass inference. To address this limitation, we propose an efficient Recursive Think–Answer Process (R-TAP) that enables models to engage in iterative reasoning cycles and generate more accurate answers, going beyond conventional single-pass approaches. Central to this approach is a confidence generator that evaluates the certainty of model responses and guides subsequent improvements. By incorporating two complementary rewards—\textit{Recursively Confidence Increase Reward} and \textit{Final Answer Confidence Reward}—we show that R-TAP-enhanced models consistently outperform conventional single-pass methods for both large language models (LLMs) and vision-language models (VLMs). Moreover, by analyzing the frequency of ``Oops''-like expressions in model responses, we find that R-TAP–applied models exhibit significantly fewer self-reflective patterns, resulting in more stable and faster inference-time reasoning. We hope R-TAP pave the way evolving into efficient and elaborated methods to refine the reasoning processes of future AI. Our project page can be found at \href{https://litcoderr.github.io/rtap_page/}{Link}.
\end{abstract}
\begin{figure*}[t]
\vspace{-8mm}
    \centering
    \includegraphics[width=0.8\linewidth]{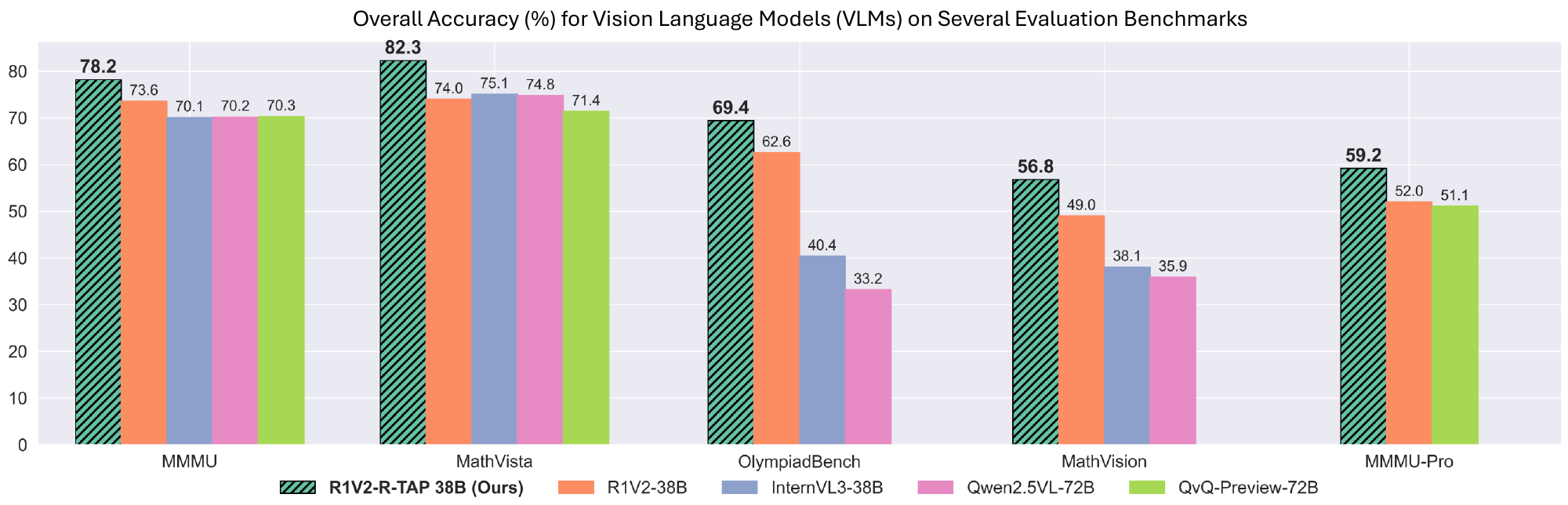}
\vspace{-3mm}
    \caption{Overall accuracy (\%) of numerous vision language models (VLMs) on five evaluation benchmarks—MMMU~\cite{yue2023mmmu}, MathVista~\cite{lu2023mathvista}, OlympiadBench~\cite{he2024olympiadbench}, MathVision~\cite{wang2024measuring}, and MMMU-Pro~\cite{yue2024mmmu}.}
    \label{fig:vlm_bench}
\vspace{-3mm}
\end{figure*}

\begin{figure*}[t]
    \vspace{0mm}
    \centering
    \includegraphics[width=0.8\linewidth]{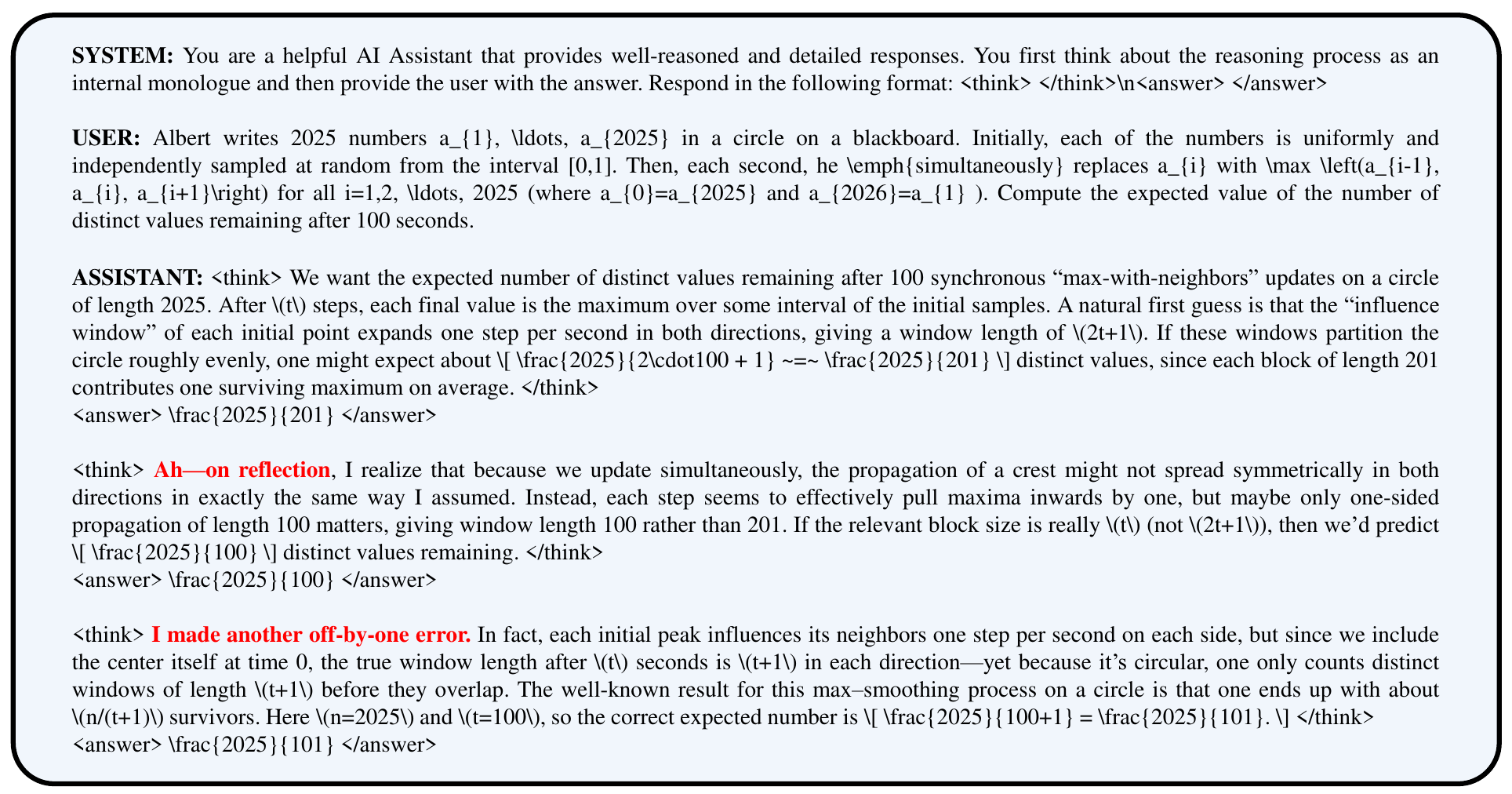}
    \vspace{-3mm}
    \caption{Qualitative example of recursive think–answer process on a combinatorics question. The model iteratively refines its solution across multiple reasoning cycles, successfully correcting initial misconceptions such as off-by-one errors.}
    \label{fig:qual}
\vspace{-3mm}
\end{figure*}
    
\section{Introduction}
\label{sec:intro}
Recent advances in a Think–Answer process-equipped models—such as OpenAI’s o1~\cite{jaech2024openai} and DeepSeek-R1~\cite{guo2025deepseek}—have demonstrated that explicitly separating the reasoning and answering stages can substantially enhance problem-solving performance. By following a Think–Answer scheme, these models surpass direct-prediction approaches and achieve remarkable capabilities in domains such as mathematical reasoning and competitive programming. This paradigm has also been extended to multimodal settings, where recent VLMs leverage a Think-Answer process~\cite{peng2025lmm, huang2025vision, chen2025r1v, deng2025openvlthinker, yang2025r1, zhou2025r1, meng2024simpo} to improve vision-language reasoning.

\begin{figure*}[t]
\vspace{-8mm}
    \centering
    \includegraphics[width=0.8\linewidth]{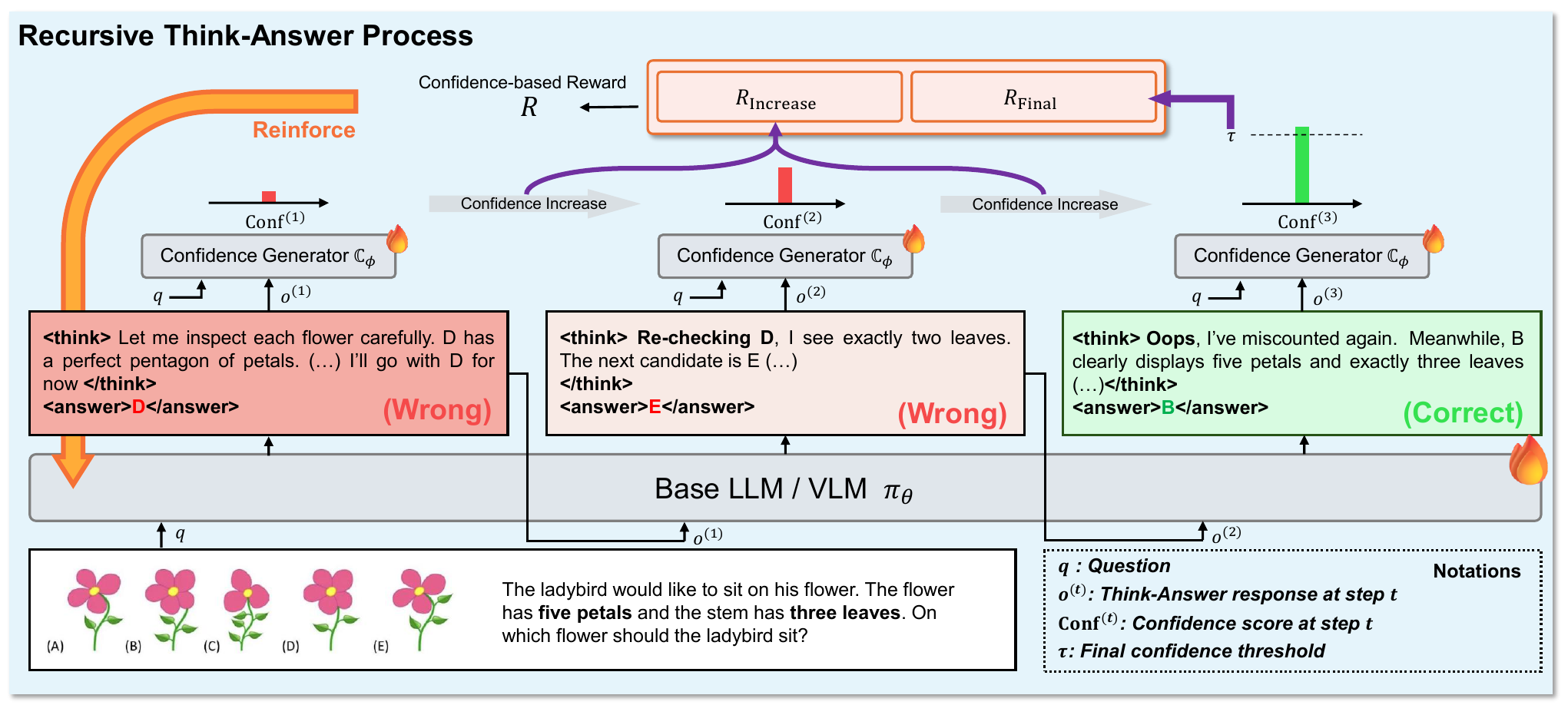}
    \vspace{-3mm}
    \caption{\textbf{Recursive Think-Answer Process.} Given a question \(q\), base LLM/VLM \(\pi_{\theta}\) recursively generates multiple Think-Answers \(o^{(t)}\) until the answer is correct $t=M$. In this example, effective recursion depth $M=\text{3}$. A pre-trained Confidence Generator $\mathbb{C}_{\phi}$ assess each question and Think-Answer pair $(q, o^{(t)})$ then generates confidence score $\text{Conf}^{(t)}$. This confidence score is used to formulate confidence-based reward -- $R_{\text{Increase}}$ and $R_{\text{Final}}$ -- which serves as a sufficient reinforcement signal to train the model to recursively generate higher confidence Think-Answers until intrinsic confidence is high enough. Note that full responses for this question is described in Appendix A.}
    \label{fig:rtap}
\vspace{-3mm}
\end{figure*}

Despite these successes, current Think–Answer models almost always rely on a single-pass reasoning trajectory. After generating one Think–Answer pair, the model stops its inference process—even when the reasoning is inaccurate, inconsistent, or clearly uncertain. Models often produce self-reflective cues such as ``Oops!'' or ``Let me try again'' which show their uncertainty. However, these signals are not used: the model outputs its final answer without any method for self-evaluation or additional refinement. As a result, incorrect but confident-sounding reasoning remains uncorrected, reducing the reliability and consistency of Think–Answer models.

This limitation mainly comes from current reinforcement learning (RL) frameworks. Recent GRPO-style~\cite{shao2024deepseekmath} sampling methods optimize only a single Think–Answer trajectory with rewards such as accuracy or format correctness. However, these methods do not consider the model’s confidence in its reasoning, so they cannot support introspective checks or recursive correction. Because of this, the model cannot tell when its answer has low confidence and needs further refinement, which becomes a serious issue for complex or high-stakes tasks.

To address these limitations, we introduce R-TAP, \textbf{R}ecursive \textbf{T}hink-\textbf{A}nswer \textbf{P}rocess that enables LLMs and VLMs to iteratively refine their reasoning through confidence-guided Think–Answer cycles. Instead of stopping after one reasoning pass, the model evaluates its own confidence after each cycle using a dedicated \textit{Confidence Generator}. When the models internally think the confidence is low, the model runs an additional Think–Answer cycle, revisiting and improving its previous reasoning. This recursive approach encourages deeper thinking, self-reflection, and step-by-step refinement of the reasoning process.

R-TAP introduces two key technical components. First is a Confidence Generator, initialized from the pretrained model and fine-tuned to quantify correctness of individual reasoning paths. Second is a recursive reward structure, combining (i) \textit{Recursively Confidence Increase Reward} that encourages confidence to improve from one cycle to the next, and (ii) \textit{Final Answer Confidence Reward} that encourages high-confidence final predictions. Together, these components provide the necessary training signals for LLMs and VLMs to learn recursive Think-Answer trajectories for strengthening their reasoning.

To implement this approach, we proceed in two stages. Stage 1 performs supervised learning on the Confidence Generator using binary correctness labels for each reasoning trajectory produced by the target model. Stage 2 applies RL with GRPO to optimize the model’s reasoning behavior under recursive rewards, enabling the model to generate progressively more accurate and confident reasoning across cycles. During this stage, the Confidence Generator is trained simultaneously to predict reliable confidence scores for the updated model’s responses in real time. Notably, the Confidence Generator is used only during training, so R-TAP introduces no additional inference-time cost. In summary, R-TAP explicitly optimizes both introspection and self-correction, allowing models to refine their own trajectories and move beyond the rigid, one-shot Think-Answer paradigm.

Our experiments show that R-TAP delivers strong and consistent performance improvements across diverse language and vision-language reasoning benchmarks. Moreover, we find that recursive refinement leads to a substantial reduction in ``Oops!''-style self-corrections during inference, indicating that R-TAP achieves more reliable yet fast inference-time reasoning with fewer failures along the trajectory. These results demonstrate that confidence-guided recursion training is a powerful mechanism for enhancing both the accuracy and inference speed of modern reasoning models. Our main contributions are summarized as follows:
\begin{itemize}
  \item \textbf{Recursive Think-Answer Process}: We propose R-TAP, a confidence-driven iterative reasoning framework that enables dynamic re-engagement of reasoning cycles and self-corrective refinement.
  \item \textbf{Unified Reasoning Across Modalities}: R-TAP generalizes effectively to both language-only and multimodal reasoning, providing a unified mechanism for recursive improvement.
\end{itemize}
\section{Related Work}
\label{sec:related}

\begin{table*}[t!]
\vspace{-8mm}
\centering
\begin{minipage}{0.49\textwidth}
\centering
\caption{Performance comparison of open-source large language models on challenging math benchmarks: AIME24~\cite{patel2024aime}, AMC~\cite{amc23}, MATH500~\cite{math500}, Minerva~\cite{minervamath}, and OlympiadBench~\cite{he2024olympiadbench}.}
\vspace{-3mm}
\label{tab:oat}
\resizebox{\textwidth}{!}{
\vspace{-3mm}
\renewcommand{\tabcolsep}{3.3mm}
\begin{tabular}{lcccccc}
\toprule
Base Model + Method & AIME24 & AMC & MATH500 & Minerva & OlympiadBench & Avg. \\
\midrule
Qwen2.5-Math-1.5B~\cite{yang2024qwen2math} & 16.7 & 43.4 & 61.8 & 15.1 & 28.4 & 33.1 \\
Qwen2.5-Math-1.5B-Instruct~\cite{yang2024qwen2math} & 10.0 & 48.2 & 74.2 & 26.5 & 40.2 & 39.8 \\
\cdashline{1-7}\noalign{\vskip 0.5ex}
R1-Distill-Qwen-1.5B @ 3k~\cite{guo2025deepseek} & 2.5 & 21.7 & 52.2 & 16.3 & 17.3 & 22.0 \\
R1-Distill-Qwen-1.5B @ 8k~\cite{guo2025deepseek} & 20.0 & 49.4 & 77.4 & 25.0 & 35.8 & 41.5 \\
\rowcolor{colorful}
\textbf{R1-Distill-Qwen-1.5B-R-TAP} @ 8k & \textbf{27.0} & \textbf{55.8} & \textbf{83.5} & \textbf{31.9} & \textbf{42.4} & \textbf{48.1} \\
\cdashline{1-7}\noalign{\vskip 0.5ex}
Oat-Zero-1.5B~\cite{liu2025there} & 20.0 & 53.0 & 74.2 & 25.7 & 37.6 & 42.1 \\
\rowcolor{colorful}
\textbf{Oat-Zero-1.5B-R-TAP} & \textbf{26.2} & \textbf{59.7} & \textbf{80.0} & \textbf{31.2} & \textbf{42.8} & \textbf{48.0} \\
\midrule
OpenReasoner-Zero-7B @ 3k~\cite{hu2025open} & 13.3 & 47.0 & 79.2 & 31.6 & 44.0 & 43.0 \\
OpenReasoner-Zero-7B @ 8k~\cite{hu2025open} & 13.3 & 54.2 & 82.4 & 31.6 & 47.9 & 45.9 \\
Qwen2.5-Math-7B~\cite{yang2024qwen2math} & 0.2 & 45.8 & 69.0 & 21.3 & 34.7 & 38.2 \\
Qwen2.5-Math-7B-Instruct~\cite{yang2024qwen2math} & 16.7 & 53.0 & 83.6 & 29.8 & 42.7 & 45.1 \\
SimpleRL-Zero-7B~\cite{zeng2025simplerl} & 26.7 & 60.2 & 78.2 & 27.6 & 40.3 & 46.6 \\
PRIME-Zero-7B~\cite{cui2025process} & 16.7 & 62.7 & 83.8 & 36.0 & 40.9 & 48.0 \\
\cdashline{1-7}\noalign{\vskip 0.5ex}
R1-Distill-Qwen-7B @ 3k~\cite{guo2025deepseek} & 10.0 & 26.2 & 60.1 & 23.0 & 23.1 & 28.5 \\
R1-Distill-Qwen-7B @ 8k~\cite{guo2025deepseek} & 33.3 & 68.4 & 88.1 & 35.9 & 47.7 & 54.7 \\
\rowcolor{colorful}
\textbf{R1-Distill-Qwen-7B-R-TAP} & \textbf{39.7} & \textbf{75.0} & \textbf{92.7} & \textbf{42.3} & \textbf{53.8} & \textbf{60.7} \\
\cdashline{1-7}\noalign{\vskip 0.5ex}
Oat-Zero-7B~\cite{liu2025there} & 43.3 & 62.7 & 80.0 & 30.1 & 41.0 & 51.4 \\
\rowcolor{colorful}
\textbf{Oat-Zero-7B-R-TAP} & \textbf{50.5} & \textbf{69.5} & \textbf{87.2} & \textbf{37.2} & \textbf{46.8} & \textbf{57.7} \\
\bottomrule
\end{tabular}
}
\end{minipage}
\hfill
\begin{minipage}{0.49\textwidth}
\centering
\caption{Comparison of closed and open-source language models with MiMo-RL-7B-R-TAB across general knowledge, mathematical, and code benchmarks. General: GPQA Diamond~\cite{rein2024gpqa}, SuperGPQA~\cite{du2025supergpqa}, DROP~\cite{dua2019drop}, MMLU-Pro~\cite{wang2024mmlu}, IF-Eval~\cite{zhou2023instruction}; Math: MATH500~\cite{math500}, AIME2024~\cite{patel2024aime}, AIME2025~\cite{patel2024aime}; Code: LiveCodeBench v5, v6~\cite{jain2024livecodebench}.}
\vspace{-3mm}
\label{tab:mimo}
\resizebox{\textwidth}{!}{
\renewcommand{\tabcolsep}{1mm}
\begin{tabular}{lccccccc>{\columncolor{colorful}}c}
\toprule
Benchmark & \makecell{GPT-4o\\0513~\cite{hurst2024gpt}} & \makecell{Claude-3.5-\\Sonnet-1022~\cite{claude3series2024}} & \makecell{OpenAI\\o1-mini~\cite{jaech2024openai}} & \makecell{QwQ-32B\\Preview~\cite{qwq32b}} & \makecell{R1-Distill-\\Qwen-14B~\cite{guo2025deepseek}} & \makecell{R1-Distill-\\Qwen-7B~\cite{guo2025deepseek}} & \makecell{MiMo-\\7B-RL~\cite{coreteam2025mimounlockingreasoningpotential}} & \makecell{\textbf{MiMo-RL}\\\textbf{7B-R-TAP}} \\
\midrule
\multicolumn{9}{c}{\textit{General}} \\
\midrule
GPQA Diamond (Pass@1) & 49.9 & \textbf{65.0} & 60.0 & 54.5 & 59.1 & 49.1 & 54.4 & 60.7 \\
SuperGPQA (Pass@1) & 42.4 & \textbf{48.2} & 45.2 & 43.6 & 40.6 & 28.9 & 40.5 & 47.3 \\
DROP (3-shot F1) & 83.7 & \textbf{88.3} & 83.9 & 71.2 & 85.5 & 77.0 & 78.7 & 84.5 \\
MMLU-Pro (EM) & 72.6 & 78.0 & \textbf{80.3} & 52.0 & 68.8 & 53.5 & 58.6 & 65.9 \\
IF-Eval (Prompt Strict) & 84.3 & \textbf{86.5} & 84.8 & 40.4 & 78.3 & 60.5 & 61.0 & 68.0 \\
\midrule
\multicolumn{9}{c}{\textit{Mathematics}} \\
\midrule
MATH500 (Pass@1) & 74.6 & 78.3 & 90.0 & 90.6 & 93.9 & 92.8 & 95.8 & \textbf{97.3} \\
AIME 2024 (Pass@1) & 9.3 & 16.0 & 63.6 & 50.0 & 69.7 & 55.5 & 68.2 & \textbf{75.8} \\
AIME 2025 (Pass@1) & 11.6 & 7.4 & 50.7 & 32.4 & 48.2 & 38.8 & 55.4 & \textbf{61.9} \\
\midrule
\multicolumn{9}{c}{\textit{Code}} \\
\midrule
LiveCodeBench v5 (Pass@1) & 32.9 & 38.9 & 53.8 & 41.9 & 53.1 & 37.6 & 57.8 & \textbf{64.2} \\
LiveCodeBench v6 (Pass@1) & 30.9 & 37.2 & 46.8 & 39.1 & 31.9 & 23.9 & 49.3 & \textbf{56.3} \\
\bottomrule
\end{tabular}
}
\end{minipage}
\vspace{-3mm}
\end{table*}

\begin{figure*}[t!]
\vspace{0mm}
    \centering
    \includegraphics[width=0.8\textwidth]{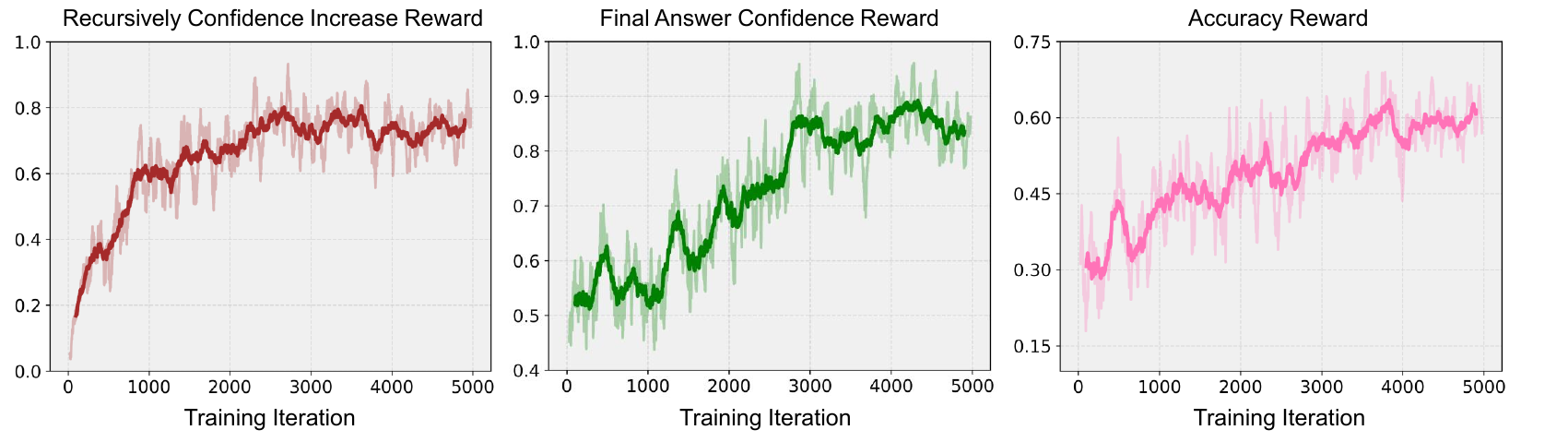}
    \vspace{-3mm}
    \caption{Training curves showing the progression of three reward signals—recursively confidence increase reward, last answer’s confidence reward, and accuracy reward—over iterations during GRPO~\cite{shao2024deepseekmath}. All rewards show consistent upward trends, indicating effective recursive refinement.}
    \label{fig:rew_curve}
\vspace{-3mm}
\end{figure*}

\vspace{1.5mm}
\noindent\textbf{Evolution of LLMs and VLMs.}
The rapid scaling of LLMs and VLMs has enabled impressive improvements in reasoning, alignment, and multimodal understanding. Early efforts such as GPT-3~\cite{brown2020language} demonstrated strong in-context learning capabilities, while subsequent alignment-oriented approaches including InstructGPT~\cite{ouyang2022training} and ChatGPT leveraged RL from human feedback (RLHF) to improve reliability and adherence to instructions. More advanced closed-source models such as GPT-4~\cite{achiam2023gpt} expanded these capabilities to multimodal settings, showcasing strong performance across diverse problem-solving tasks.

Open-source research has paralleled these advances, driven by architectural innovations and large-scale instruction tuning. LLaMA~\cite{touvron2023llama} and its successors introduced lightweight yet performant architectures, while models such as LLaVA-NeXT~\cite{liu2024llavanext}, MM1~\cite{mckinzie2024mm1}, Yi-VL~\cite{young2024yi}, and MiniGemini~\cite{li2024mini} scaled VLM reasoning by increasing training data and integrating powerful vision encoders. Additional efforts incorporate specialized projectors or structured modules~\cite{tong2024cambrian, ge2024convllava, chen2024evlm, yao2024minicpm} to extract hierarchical or task-specific features. Despite these advancements, most models still depend on single-pass predictions at inference time, without iterative introspection or self-correction.

\vspace{1.5mm}
\noindent\textbf{Think-Answer Reasoning.}
Think–Answer paradigm has emerged as a compelling approach for enhancing reasoning in both LLMs and VLMs. Chain-of-Thought prompting~\cite{wei2022chain} first demonstrated the benefits of eliciting intermediate reasoning steps, followed by extensions such as Program-of-Thoughts~\cite{chen2022program}, Tree of Thoughts~\cite{yao2023tree}, and Graph of Thoughts~\cite{besta2024graph} that explore larger reasoning spaces through structured search. While these methods improve robustness, they typically rely on external sampling or reranking rather than intrinsic self-evaluation.

Several reinforcement-learning-based methods introduce iterative refinement signals, including Reflexion~\cite{shinn2023reflexion} and Self-Consistency~\cite{wang2022self}, but they operate outside the Think–Answer framework or depend on majority voting instead of introspective certainty. Recent Think-Answer models—DeepSeek-R1~\cite{guo2025deepseek} and OpenAI o1~\cite{jaech2024openai}—demonstrate that separating “thinking” from “answering” can yield strong results in mathematical reasoning and programming. Similar trends are observed in multimodal reasoning, where models such as LMM-R1~\cite{peng2025lmm}, Vision-R1~\cite{huang2025vision}, R1-V~\cite{chen2025r1v}, and R1-Zero~\cite{zhou2025r1} adopt slow-thinking trajectories for visual problem solving.

However, almost all existing Think–Answer systems adopt a \emph{single-pass} reasoning trajectory: after producing one Think–Answer pair, the model terminates inference even when its reasoning is uncertain or contains explicit self-reflective cues (e.g., ``Oops!'' and ``Let me try again''). Sampling-based re-ranking strategies partially mitigate this issue but do not enable the model to internally assess confidence or decide whether further reasoning is needed. 

\vspace{1.5mm}
\noindent\textbf{Recursive and Confidence-Guided Refinement.}
Although iterative refinement has been explored through external verification~\cite{wang2022self} or heuristic feedback~\cite{shinn2023reflexion}, prior works lack a principled mechanism to internally estimate the correctness of reasoning and dynamically invoke additional reasoning cycles. No existing Think–Answer model explicitly incorporates a confidence predictor to guide recursive inference-time reasoning.

\begin{table*}[t!]
\vspace{-5mm}
\centering
\begin{minipage}{0.49\textwidth}
\centering
\vspace{-3mm}
\caption{Performance comparison across various models including our proposed AZR variants on coding and mathematical reasoning benchmarks: HumanEval~\cite{chen2021codex}, MBPP~\cite{austin2021program}, LiveCodeBench (LCB)~\cite{austin2021program}, AIME~\cite{patel2024aime}, AMC~\cite{amc23}, MATH500~\cite{math500}, Minerva~\cite{minervamath}, and OlympiadBench~\cite{he2024olympiadbench}.}
\vspace{-3mm}
\label{tab:azr}
\resizebox{\textwidth}{!}{
\renewcommand{\tabcolsep}{0.8mm}
\begin{tabular}{lcccccccccccccc}
\toprule
Model & Base 
& HEval & MBPP & LCB 
& AME24 & AME25 & AMC & M500 
& Minerva & Olympiad 
& CAvg & MAvg & AVG \\
\midrule
\multicolumn{15}{c}{\textit{Base Models}} \\
\midrule
Qwen2.5-7B~\cite{yang2024qwen2} & - & 73.2 & 65.3 & 17.5 & 6.7 & 3.3 & 37.5 & 64.8 & 25.0 & 27.7 & 52.0 & 27.5 & 39.8 \\
Qwen2.5-7B-Ins~\cite{yang2024qwen2} & - & 75.0 & 68.5 & 25.5 & 13.3 & 6.7 & 52.5 & 76.4 & 35.7 & 37.6 & 56.3 & 37.0 & 46.7 \\
Qwen2.5-7B-Coder~\cite{yang2024qwen2} & Coder & 80.5 & 69.3 & 19.9 & 13.3 & 6.7 & 40.0 & 54.0 & 17.1 & 21.9 & 56.6 & 33.9 & 40.2 \\
Qwen2.5-7B-Math~\cite{yang2024qwen2} & - & 61.0 & 57.9 & 16.2 & 10.0 & 16.7 & 42.5 & 64.2 & 15.4 & 28.0 & 45.0 & 29.5 & 37.3 \\
\midrule
\multicolumn{15}{c}{\textit{Zero-Style Reasoners for Code}} \\
\midrule
AceCoder-RM~\cite{zeng2025acecoder} & Ins & 79.9 & 71.4 & 23.6 & 20.0 & 6.7 & 50.0 & 76.4 & 34.6 & 36.7 & 58.3 & 37.4 & 47.9 \\
AceCoder-Rule~\cite{zeng2025acecoder} & Ins & 77.4 & 69.0 & 19.9 & 13.3 & 6.7 & 50.0 & 76.0 & 37.5 & 37.8 & 55.4 & 36.9 & 46.2 \\
AceCoder-RM~\cite{zeng2025acecoder} & Coder & 78.0 & 66.4 & 27.5 & 13.3 & 6.7 & 50.0 & 62.6 & 29.4 & 29.0 & 57.3 & 27.5 & 42.4 \\
AceCoder-Rule~\cite{zeng2025acecoder} & Coder & 80.5 & 70.4 & 29.0 & 6.7 & 6.7 & 37.5 & 62.8 & 27.6 & 27.4 & 60.0 & 28.5 & 44.8 \\
CodeR1-LC2k~\cite{code-r1} & Ins & 81.7 & 71.7 & 28.1 & 13.3 & 3.3 & 40.0 & 74.0 & 35.6 & 36.7 & 60.3 & 35.8 & 48.6 \\
CodeR1-12k~\cite{code-r1} & Ins & 81.1 & 73.5 & 29.9 & 13.3 & 3.3 & 37.5 & 74.0 & 35.7 & 36.9 & 61.3 & 33.5 & 47.4 \\
PRIME-Zero~\cite{cui2025process} & Coder & 49.4 & 51.1 & 11.0 & 23.3 & \textbf{23.3} & \textbf{67.5} & \textbf{81.2} & 37.9 & 41.8 & 37.2 & 45.8 & 41.5 \\
AZR~\cite{zhao2025absolute}  & Coder & 83.5 & 69.6 & 31.7 & 20.0 & 10.0 & 57.5 & 72.6 & 36.4 & 38.2 & 61.6 & 39.1 & 50.4 \\
\rowcolor{colorful}
\textbf{AZR-R-TAP} & Coder & \textbf{89.3} & \textbf{75.9} & \textbf{38.2} & \textbf{27.4} & 18.2 & 64.9 & 78.1 & \textbf{44.3} & \textbf{46.6} & \textbf{67.8} & \textbf{46.6} & \textbf{57.2} \\
\midrule
\multicolumn{15}{c}{\textit{Zero-Style Reasoners for Math}} \\
\midrule
SimpleRL-Zoo~\cite{zeng2025simplerl} & Base & 62.3 & 63.2 & 25.5 & \textbf{20.0} & 3.3 & \textbf{67.5} & 80.7 & 37.1 & 41.0 & 54.0 & 38.5 & 46.5 \\
Oat-Zero~\cite{liu2025there} & Math & 62.2 & 59.0 & 15.2 & 16.7 & 3.3 & 66.7 & 65.5 & 44.3 & 44.9 & 51.6 & 35.4 & 43.4 \\
ORZ~\cite{hu2025openreasonerzeroopensourceapproach} & Base & 80.5 & 64.3 & 22.0 & 13.3 & 16.7 & 60.0 & 81.8 & 32.7 & \textbf{45.0} & 55.6 & \textbf{41.6} & 48.6 \\
AZR~\cite{zhao2025absolute} & Base & 71.3 & 69.1 & 25.3 & 13.3 & 13.3 & 52.5 & 74.4 & 38.2 & 38.2 & 55.2 & 39.0 & 46.8 \\
\rowcolor{colorful}
\textbf{AZR-R-TAP} & Base & \textbf{78.5} & \textbf{76.7} & \textbf{32.4} & 19.0 & \textbf{20.1} & 59.8 & \textbf{81.2} & \textbf{45.7} & 44.4 & \textbf{62.5} & 38.4 & \textbf{50.5} \\
\bottomrule
\end{tabular}

}
\end{minipage}
\hfill
\begin{minipage}{0.49\textwidth}
\centering
\vspace{-3mm}
\caption{Comparison of mathematical and coding benchmark performance across various instruction-tuned reinforcement-tuned and R-TAP integrated models. Benchmarks include AIME~\cite{patel2024aime}, AMC~\cite{amc23}, MATH500~\cite{math500}, MinervaMath~\cite{minervamath}, OlympiadBench~\cite{he2024olympiadbench}, LeetCodeBench~\cite{coignion2024performance}, and LiveCodeBench~\cite{jain2024livecodebench}.}
\vspace{-3mm}
\label{tab:prime}
\resizebox{\textwidth}{!}{
\renewcommand{\tabcolsep}{1mm}
\begin{tabular}{lcccccccc}
\toprule
Method & AIME 2024 & AMC & MATH500 & MinervaMath & OlympiadBench & LeetCodeBench & LiveCodeBench & Avg. \\
\midrule
GPT-4o~\cite{hurst2024gpt} & 9.3 & 45.8 & 76.4 & 36.8 & 43.3 & \textbf{58.9} & \textbf{48.8} & 45.6 \\
Llama-3.1-70B-Inst.~\cite{grattafiori2024llama} & 20.0 & 37.3 & 65.0 & 37.1 & 30.5 & 35.0 & 34.4 & 37.0 \\
Qwen2.5-Math-7B-Inst.~\cite{yang2024qwen2} & 13.3 & 50.6 & 79.8 & 34.6 & 40.7 & 11.7 & 11.3 & 34.6 \\
Eurus-2-7B-SFT~\cite{yuan2024advancing} & 3.3 & 30.1 & 66.2 & 32.7 & 29.8 & 21.7 & 17.8 & 28.8 \\
RLOO~\cite{ahmadian2024back} & 20.0 & 47.0 & 73.2 & 36.4 & 35.4 & 28.3 & 26.7 & 36.9 \\
Eurus-2-7B-PRIME~\cite{yuan2024advancing} & 20.0 & 50.6 & 78.2 & 39.3 & 40.3 & 31.1 & 27.5 & 41.0 \\
\rowcolor{colorful}
\textbf{Eurus-2-7B-PRIME-R-TAP} & \textbf{28.3} & \textbf{57.5} & \textbf{83.5} & \textbf{43.8} & \textbf{47.4} & 38.6 & 31.8 & \textbf{47.2} \\
\bottomrule
\end{tabular}

}

\centering
\caption{GRPO-trained models on MathVerse~\cite{zhang2024mathverse}, MathVision~\cite{wang2024measuring}, MathVista~\cite{lu2023mathvista}, WeMath~\cite{qiao2024we}, and HallusionBench~\cite{liu2023hallusionbench}.}
\vspace{-3mm}
\label{tab:noisy}
\resizebox{\textwidth}{!}{
\renewcommand{\tabcolsep}{5mm}
\begin{tabular}{lcccccc}
\toprule
Model & MathVerse & MathVision & MathVista & WeMath & HallusionBench & Avg. \\
\midrule
R1-VL-7B~\cite{zhang2025r1} & 40.0 & 24.7 & 63.5 & - & - & - \\
Vision-R1-7B~\cite{huang2025vision} & 52.4 & - & 73.5 & - & - & - \\
R1-OneVision-7B~\cite{yang2025r1} & 46.1 & 22.5 & 63.9 & 62.1 & 65.6 & 52.0 \\
OpenVLThinker-7B~\cite{deng2025openvlthinker} & 48.0 & 25.0 & 71.5 & 67.8 & 70.8 & 56.5 \\
MM-Eureka-Qwen-7B~\cite{meng2025mm} & 50.5 & 28.3 & 71.5 & 65.5 & 68.3 & 56.8 \\
ADORA-7B~\cite{gui2025adora} & 50.1 & 27.6 & 71.1 & 67.1 & 53.1 & 53.8 \\
ThinkLite-7B-VL~\cite{wang2025sota} & 50.2 & 27.6 & 72.7 & 69.2 & 71.0 & 58.1 \\
VLAA-Thinker-Qwen2.5-7B~\cite{chen2025sft} & 49.9 & 26.9 & 68.8 & 67.9 & 68.6 & 56.4 \\
\cdashline{1-7}\noalign{\vskip 0.5ex}
Qwen2.5-VL-7B-Instruct~\cite{bai2025qwen2} & 46.2 & 25.0 & 67.5 & 63.1 & 71.2 & 53.3 \\
\quad + Vanilla GRPO~\cite{shao2024deepseekmath} & 50.7 & 28.5 & 71.7 & 68.6 & 69.8 & 57.9 \\
\quad + NoisyRollout~\cite{liu2025noisyrollout} & 52.8 & 28.9 & 72.9 & 71.9 & 70.8 & 59.5 \\
\rowcolor{colorful}
\quad + \textbf{R-TAP} & \textbf{60.1} & \textbf{35.3} & \textbf{79.4} & \textbf{78.2} & \textbf{77.9} & \textbf{66.2} \\
\bottomrule
\end{tabular}

}

\end{minipage}
\vspace{-3mm}
\end{table*}

In contrast, R-TAP introduces a confidence-aware recursive Think–Answer process that enables models to introspect, detect low-confidence reasoning, and selectively re-engage in additional reasoning cycles. By measuring both the recursive increase in confidence across reasoning cycles and the confidence of the final prediction, R-TAP offers a unified and efficient mechanism for self-corrective reasoning in both LLMs and VLMs, moving beyond the static, single-pass nature of prior approaches.
\section{R-TAP: Recursive Think--Answer Process}
\label{sec:method}

R-TAP enables a model to perform confidence-guided recursive reasoning, allowing it to 
(i) evaluate the reliability of its reasoning, 
(ii) continue reasoning when uncertain, and 
(iii) terminate early when sufficiently confident. 
This section presents the overall framework, the Confidence Generator, the recursive generation scheme, 
the confidence-based reward design, and implementation details.

\subsection{Problem Definition and Overall Framework}
\label{subsec:framework}

Given a question $q$, conventional single-step reasoning LLMs/VLMs produce exactly single Think-Answer process:
\begin{equation}
    o^{(1)} \sim \pi_{\theta}(o\mid q),
\end{equation}
and terminate immediately, even when the produced reasoning is uncertain or incorrect. 
Such models lack an internal mechanism for introspection or self-initiated refinement.

R-TAP generalizes this paradigm by allowing the model to recursively generate recursive 
Think-Answer responses $\mathcal{O}$ as follows:
\begin{equation}
\begin{aligned}
\mathcal{O} = \{&o^{(1)}, o^{(2)}, \ldots, o^{(T)}\}\sim\pi_{\theta}(\mathcal{O}\mid q),\\\\
\text{such that}&\quad o^{(t+1)} \sim \pi_{\theta}(o\mid q, \{o^{(i)}\}_{i=1}^{t}),
\end{aligned}
\label{eq:rtap-out}
\end{equation}
where $T$ denotes the recursion depths. During training, $T$ is fixed to do efficient batch sampling, while at inference time the model internally determines whether to continue or terminate.

\begin{table*}[t!]
\vspace{-5mm}
\centering
\begin{minipage}{0.49\textwidth}
\centering
\vspace{-3mm}
\caption{Performance of closed-source and open-source multimodal models on the R1-OneVision benchmark~\cite{yang2025r1}. The benchmark is organized by diverse education level and subject category.}
\vspace{-3mm}
\label{tab:r1ov}
\resizebox{\textwidth}{!}{
\renewcommand{\tabcolsep}{1mm}
\begin{tabular}{lcccccccccccc}
\toprule
\multirow{2}{*}{Model} & \multirow{2}{*}{Avg} & \multicolumn{4}{c}{Grade} & \multicolumn{5}{c}{Category} \\
& & Junior High School & High School & College & Social Test & Math & Physics & Chemistry & Biology & Deduction \\
\midrule
\multicolumn{12}{c}{\textit{Closed-source}} \\
\midrule
GPT-4o~\cite{hurst2024gpt} & 49.6 & 51.3 & 56.2 & 45.3 & 26.5 & 41.3 & 52.5 & 71.4 & 63.4 & 26.5 \\
Gemini-2.0-Flash~\cite{team2023gemini} & 59.1 & 56.0 & 65.9 & 61.2 & 39.8 & 52.3 & 64.4 & 74.3 & 67.2 & 39.8 \\
Claude-3.5~\cite{claude3series2024} & 52.1 & 56.0 & 55.9 & 49.4 & 30.6 & 46.5 & 54.3 & 66.7 & 65.7 & 30.6 \\
\midrule
\multicolumn{12}{c}{\textit{Open-source}} \\
\midrule
MiniCPM-o-2.6~\cite{hu2024minicpm} & 30.4 & 33.4 & 31.7 & 21.2 & 31.6 & 24.2 & 31.7 & 30.5 & 41.8 & 31.6 \\
InternVL2.5-8B~\cite{chen2024expanding} & 29.5 & 33.1 & 30.6 & 21.8 & 27.6 & 26.3 & 24.8 & 32.4 & 46.3 & 27.6 \\
InternVL2.5-8B-MPO~\cite{chen2024expanding} & 32.5 & 37.4 & 33.6 & 24.7 & 26.5 & 28.7 & 29.9 & 41.0 & 44.8 & 26.5 \\
Qwen2-VL-7B~\cite{wang2024qwen2vl} & 30.0 & 35.4 & 32.5 & 25.9 & 26.5 & 26.3 & 28.1 & 30.5 & 45.5 & 26.5 \\
Qwen2.5-VL-7B~\cite{wang2024qwen2vl} & 32.1 & 35.3 & 37.1 & 25.9 & 19.4 & 31.5 & 27.3 & 39.0 & 47.0 & 19.4 \\
DeepSeek-VL2~\cite{wu2024deepseekvl2mixtureofexpertsvisionlanguagemodels} & 29.8 & 34.4 & 30.9 & 18.8 & 30.6 & 23.5 & 28.4 & 29.5 & 47.8 & 30.6 \\
R1-Onevision-7B~\cite{yang2025r1} & 36.2 & 40.1 & 39.5 & 27.6 & 26.5 & 33.0 & 30.2 & 49.5 & 53.0 & 26.5 \\
\rowcolor{colorful}
\textbf{R1-Onevision-7B-R-TAP} & \textbf{42.4} & \textbf{47.4} & \textbf{45.0} & \textbf{34.9} & \textbf{31.4} & \textbf{39.4} & \textbf{36.9} & \textbf{55.9} & \textbf{59.6} & \textbf{31.2} \\
\midrule
Qwen2.5-VL-72B~\cite{bai2025qwen2} & 52.0 & 54.3 & 56.7 & 54.1 & 23.5 & 48.9 & 55.8 & 63.8 & 63.4 & 23.5 \\
\bottomrule
\end{tabular}

}

\centering
\caption{Comparing the performances on AIME2024~\cite{patel2024aime}, AIME2025~\cite{patel2024aime}, LiveCodeBench~\cite{jain2024livecodebench}, and Arena-Head~\cite{arenahard2024}.}
\vspace{-3mm}
\label{tab:amthink}
\resizebox{\textwidth}{!}{
\renewcommand{\tabcolsep}{1mm}
\begin{tabular}{lcccc}
\toprule
Model & AIME2024 & AIME2025 & LiveCodeBench (v5, 2024.10--2025.02) & Arena-Hard \\
\midrule
AM-Thinking-v1 (Dense, 32B)~\cite{ji2025amthinkingv1advancingfrontierreasoning}                         & 85.3 & 74.4 & 70.3 & 92.5 \\
\rowcolor{colorful}
\textbf{AM-Thinking-v1-R-TAP} (Dense, 32B)          & \textbf{90.1} & \textbf{79.6} & \textbf{76.5} & \textbf{94.3} \\
\cdashline{1-5}\noalign{\vskip 0.5ex}
Qwen3-235B-A22B (MoE, 235B)~\cite{yang2025qwen3technicalreport}                         & 85.7 & 81.5 & 70.7 & 95.6 \\
\rowcolor{colorful}
\textbf{Qwen3-235B-A22B-R-TAP} (MoE, 235B)          & \textbf{90.9} & \textbf{88.7} & \textbf{77.6} & \textbf{97.1} \\
\cdashline{1-5}\noalign{\vskip 0.5ex}
Qwen3-32B (Dense, 32B)~\cite{yang2025qwen3technicalreport}                              & 81.4 & 72.9 & 65.7 & 93.8 \\
\rowcolor{colorful}
\textbf{Qwen3-32B-R-TAP} (Dense, 32B)               & \textbf{87.9} & \textbf{78.2} & \textbf{71.0} & \textbf{95.6} \\
\cdashline{1-5}\noalign{\vskip 0.5ex}
DeepSeek-R1 (MoE, 671B)~\cite{guo2025deepseek}                             & 79.8 & 70.0 & 64.3 & 93.2 \\
Nemetron-Ultra-253B (Dense, 256B)~\cite{bercovich2025llamanemotronefficientreasoningmodels}                   & 80.8 & 72.5 & 68.1 & 87.0 \\
OpenAI-o1 (2024-12-17)~\cite{jaech2024openai}                              & 74.3 & 79.2 & 63.9 & 92.1 \\
OpenAI-o3-mini (Medium)~\cite{gpto3minisyscard}                             & 79.6 & 74.8 & 66.3 & 89.0 \\
Gemini2.5-Pro~\cite{team2023gemini}                                       & 92.0 & 86.7 & 70.4 & 96.4 \\
\bottomrule
\end{tabular}
}

\end{minipage}
\hfill
\begin{minipage}{0.49\textwidth}
\centering
\vspace{-3mm}
\caption{Comparison of closed-source, general-purpose open-source, reasoning-focused, and proposed models on math-related benchmarks: MathVista~\cite{lu2023mathvista}, MathVerse~\cite{zhang2024mathverse}, MathVision~\cite{wang2024measuring}, OlympiadBench~\cite{he2024olympiadbench}, WeMath~\cite{qiao2024we}.}
\vspace{-3mm}
\label{tab:eureka}
\resizebox{\textwidth}{!}{
\renewcommand{\tabcolsep}{2.2mm}
\begin{tabular}{lccccc}
\toprule
Model & MathVista & MathVerse & MathVision & OlympiadBench & WeMath \\
\midrule
\multicolumn{6}{c}{\textit{Closed-Source Models}} \\
\midrule
Claude3.7-Sonnet~\cite{claude3series2024} & 66.8 & 52.0 & 41.3 & 48.9 & 72.6 \\
GPT-4o~\cite{hurst2024gpt} & 63.8 & 50.2 & 30.4 & 35.0 & 68.8 \\
o1~\cite{jaech2024openai} & 73.9 & 57.0 & 60.3 & 68.0 & 98.7 \\
Gemini2-flash~\cite{team2023gemini} & 70.4 & 59.3 & 41.3 & 51.0 & 71.4 \\
\midrule
\multicolumn{6}{c}{\textit{Open-Source General Models}} \\
\midrule
InternVL2.5-VL-8B~\cite{chen2024expanding} & 64.4 & 39.5 & 19.7 & 12.3 & 53.5 \\
Qwen-2.5-VL-7B~\cite{bai2025qwen2} & 68.2 & 47.9 & 25.4 & 20.2 & 62.1 \\
InternVL2.5-VL-38B~\cite{chen2024expanding} & 71.9 & 49.4 & 31.8 & 32.0 & 67.5 \\
Qwen-2.5-VL-32B~\cite{bai2025qwen2} & 74.7/71.7 & 49.9 & 40.1 & 30.0 & 69.1 \\
InternVL2.5-VL-78B~\cite{chen2024expanding} & 72.3 & 51.7 & 32.2 & 31.1 & 66.3 \\
Qwen-2.5-VL-72B~\cite{bai2025qwen2} & 74.8 & 57.6 & 38.1 & 40.4 & 72.4 \\
\midrule
\multicolumn{6}{c}{\textit{Open-Source Reasoning Models}} \\
\midrule
InternVL2.5-8B-MPO~\cite{chen2024expanding} & 68.9 & 35.5 & 21.5 & 7.8 & 53.5 \\
InternVL2.5-38B-MPO~\cite{chen2024expanding} & 73.8 & 46.5 & 32.3 & 25.6 & 66.2 \\
QVQ-72B-Preview~\cite{team2023gemini} & 71.4 & 48.2 & 35.9 & 33.2 & 65.4 \\
ADORA-7B~\cite{gui2025adora} & 73.5 & 50.1 & 23.0 & 20.1 & 64.2 \\
R1-Onevision-7B~\cite{yang2025r1} & 64.1 & 47.1 & 29.9 & 17.3 & 61.8 \\
OpenVLThinker-7B~\cite{deng2025openvlthinker} & 70.2 & 47.9 & 25.3 & 20.1 & 64.3 \\
\cdashline{1-6}\noalign{\vskip 0.5ex}
MM-Eureka-7B~\cite{meng2025mm} & 73.0 & 50.3 & 26.9 & 20.1 & 66.1 \\
\rowcolor{colorful}
\textbf{MM-Eureka-7B-R-TAP} & \textbf{79.3} & \textbf{56.1} & \textbf{31.7} & \textbf{27.5} & \textbf{71.0} \\
\cdashline{1-6}\noalign{\vskip 0.5ex}
MM-Eureka-32B~\cite{meng2025mm} & 74.8 & 56.5 & 34.4 & 35.9 & 73.4 \\
\rowcolor{colorful}
\textbf{MM-Eureka-32B-R-TAP} & \textbf{80.2} & \textbf{61.8} & \textbf{39.9} & \textbf{41.2} & \textbf{79.3} \\
\bottomrule
\end{tabular}

}
\end{minipage}
\vspace{-3mm}
\end{table*}

Similar to prior single-pass Think-Answer models~~\cite{guo2025deepseek}, R-TAP also trains the model by maximizing the objective of GRPO~\cite{shao2024deepseekmath} based on $G$ generated samples, which can be written as follows:
\begin{equation}
\begin{aligned}
    &\max_{\theta}\mathbb{E}_{q\sim\mathcal{D}, \{\mathcal{O}_{i}\}_{i=1}^{G}\sim\pi_{\theta}(\mathcal{O}\mid q)}[\mathbb{E}_{i}[ \\
    &\min\left(r_i A_i,\text{clip}(r_i, 1-\epsilon, 1+\epsilon) A_i\right)-\beta\mathcal{D}_{\mathrm{KL}}(\pi_{\theta}\|\pi_{\mathrm{ref}})]],
\end{aligned}
\label{eq:grpo}
\end{equation}
where the policy ratio $r_i$ and advantage $A_i$ are defined by
\begin{equation}
r_i = \frac{\pi_{\theta}(\mathrm{O}_i \mid q)}{\pi_{\theta_{\mathrm{old}}}(\mathrm{O}_i \mid q)},
\qquad
A_i = \frac{R_i - \mathbb{E}[\{R_i\}_{i=1}^{G}]}{\sigma[\{R_i\}_{i=1}^{G}]},
\end{equation}
with $R_i$ denotes the total reward for recursive trajectory $\mathcal{O}_i$ which will be explained in next section. This objective encourages the model to engage in further recursive reasoning when it internally detects low confidence, while allowing it to stop once its confidence reaches a satisfactory level.

\subsection{Confidence Generator}
\label{subsec:conf_gen}

A central component of R-TAP is the \emph{Confidence Generator} $\mathbb{C}_{\phi}$, 
which provides the reliability of each response in recursive Think-Answer trajectory.  
Importantly, $\mathbb{C}_{\phi}$ is used only during R-TAP training and removed at inference, so R-TAP introduces no inference-time cost.

\paragraph{Architecture.}
Given a question $q$ and one of Think-Answer responses: $o^{(t)}$ in \cref{eq:rtap-out},  
the Confidence Generator $\mathbb{C}_{\phi}$ outputs a scalar confidence score from zero to one continuous value:
\begin{equation}
\text{Conf}^{(t)} = \mathbb{C}_{\phi}(q, o^{(t)}).
\end{equation}
We build Confidence Generator from $\pi_{\text{ref}}$ structure in the beginning but replace the language head with a confidence head $\mathbb{R}^{d \times 1}$ ($d$: hidden dimension), followed by a sigmoid activation.

\paragraph{Supervised pre-training.}
Prior to R-TAP training, we train $\mathbb{C}_{\phi}$ to predict confidence of the given response. To implement it, we utilize the binary classification for whether the response  is correct compared with ground-truth label. For each question $q$, the model generates $N$ single Think-Answer process samples.
Let $\{o_i^{\mathrm{correct}}\}_{i=1}^{K}$ denote samples with correct answers and 
$\{o_i^{\mathrm{wrong}}\}_{i=1}^{N-K}$ denote incorrect ones.  
We optimize $\mathbb{C}_{\phi}$ using binary classification objective:
\begin{equation}
\begin{aligned}
\max_{\phi}&[\frac{1}{K}\sum_{i=1}^{K}\log \mathbb{C}_{\phi}(q,o_i^{\mathrm{correct}}) \\
&+ \frac{1}{N-K}\sum_{i=1}^{N-K}\log\!\big(1 - \mathbb{C}_{\phi}(q,o_i^{\mathrm{wrong}})\big)],
\end{aligned}
\label{eq:conf_train}
\end{equation}
where we technically set $K \approx N/2$ for balanced training. This step equips $\mathbb{C}_{\phi}$ with the ability to evaluate the reliability of recursive Think-Answer trajectories.

\subsection{Confidence-Based Reward Design}
\label{subsec:rew_design}

R-TAP introduces two complementary confidence-driven rewards that (1) encourage refinement when necessary and (2) encourage termination when sufficiently confident.

\paragraph{Recursive confidence increase reward.}
To reward meaningful refinement across recursive steps, we define:
\begin{equation}
R_{\mathrm{Increase}} = \frac{1}{M-1}\sum_{t=1}^{M-1}\mathbbm{1}\left[\text{Conf}^{(t+1)} > \text{Conf}^{(t)}\right],
\end{equation}
where $M$ is the effective recursion depth. For example, if the third Think-Answer response is correct but the previous ones are not, then $M=3$ is satisfied, but if the first Think-Answer response is correct, because $M=1$, we define this case to one reward value. Note that $M$ cannot be larger than $T$ such that $M\leq T$ because we pre-define $T$ for effective batch sampling to generate recursive Think-Answer responses. 

\paragraph{Final answer confidence reward.}
The final answer must be sufficiently confident:
\begin{equation}
R_{\mathrm{Final}} = \mathbbm{1}\left[\text{Conf}^{(M)} \ge \tau\right],
\end{equation}
where $\tau$ is a preset threshold.

\paragraph{Combined reward.}
The total reward is simply calculated by the addition, which is written as follows
\begin{equation}
R
= R_{\text{Increase}}
+ R_{\text{Final}}
+ R_{\text{Format}}
+ R_{\text{Answer}}
+ R_{\text{Length}},
\label{eq:total_reward}
\end{equation}
where we equally use the conventional rewards used in Open-R1~\cite{openr1}.
$R_{\text{Format}}$ is the reward for Think-Answer format,  
$R_{\text{Answer}}$ is for model response' correctness, and
$R_{\text{Length}}$ is for a soft length penalty. Appendix B describes the algorithms for R-TAP.

\begin{table*}[t!]
\vspace{-8mm}
\centering
\caption{Performance of Llama3.1~\cite{grattafiori2024llama}, DeepSeek-Math~\cite{shao2024deepseekmath}, Mistral~\cite{jiang2024identifying}, and Qwen-2.5~\cite{yang2024qwen2} models before and after SimpleRL-Zoo~\cite{zeng2025simplerl} fine-tuning and R-TAP integration across various mathematical reasoning benchmarks. Benchmarks include GSM8K~\cite{cobbe2021gsm8k}, MATH500~\cite{math500}, Minerva~\cite{minervamath}, OlympiadBench~\cite{he2024olympiadbench}, AIME~\cite{patel2024aime}, and AMC~\cite{amc23}. Note that, the selection of evaluation benchmarks and their performance results, and the selection of the compared models are referred to the manuscript of SimpleRL-Zoo~\cite{zeng2025simplerl}.}
\vspace{-3mm}
\label{tab:simp}
\renewcommand{\tabcolsep}{8mm}
\resizebox{0.8\textwidth}{!}{%
\begin{tabular}{lcccccccc}
\toprule
Model & GSM8K & MATH500 & Minerva Math & Olympiad Bench & \makecell{AIME24\\(P@1)} & \makecell{AIME24\\(Avg@32)} & AMC23 \\
\midrule
\multicolumn{9}{c}{\textit{Llama, DeepSeek and Mistral Models}} \\
\midrule
Mistral-v0.1-7B~\cite{jiang2024identifying} & 21.2 & 4.2 & 4.0 & 2.4 & 0.0 & 0.0 & 0.0 \\
+ SimpleRL-Zoo & 75.0 & 15.8 & 6.6 & 4.1 & 0.0 & 0.2 & 10.0 \\
\rowcolor{colorful}
+ \textbf{R-TAP} & \textbf{81.2} & \textbf{21.3} & \textbf{13.9} & \textbf{10.0} & \textbf{6.4} & \textbf{9.5} & \textbf{17.4} \\
\cdashline{1-9}\noalign{\vskip 0.5ex}
Llama-3.1-8B~\cite{dubey2024llama} & 39.7 & 13.6 & 4.8 & 3.1 & 0.0 & 0.2 & 2.5 \\
+ SimpleRL-Zoo & 79.2 & 23.0 & 9.6 & 5.3 & 0.0 & 0.2 & 15.0 \\
\rowcolor{colorful}
+ \textbf{R-TAP} & \textbf{86.5} & \textbf{30.4} & \textbf{16.3} & \textbf{13.7} & \textbf{6.7} & \textbf{7.9} & \textbf{22.3} \\
\cdashline{1-9}\noalign{\vskip 0.5ex}
DeepSeek-Math-7B~\cite{shao2024deepseekmath} & 28.4 & 19.4 & 5.5 & 4.7 & 0.0 & 0.0 & 10.0 \\
+ SimpleRL-Zoo & 78.5 & 39.6 & 21.0 & 12.6 & 3.3 & 0.6 & 20.0 \\
\rowcolor{colorful}
+ \textbf{R-TAP} & \textbf{83.7} & \textbf{45.2} & \textbf{28.4} & \textbf{20.1} & \textbf{10.2} & \textbf{8.2} & \textbf{27.5} \\
\cdashline{1-9}\noalign{\vskip 0.5ex}
Mistral-Small-24B~\cite{jiang2024identifying} & 78.6 & 43.6 & 10.7 & 11.6 & 3.3 & 0.5 & 17.5 \\
+ SimpleRL-Zoo & 92.0 & 70.6 & 36.8 & 36.6 & 16.7 & 13.1 & 45.0 \\
\rowcolor{colorful}
+ \textbf{R-TAP} & \textbf{95.4} & \textbf{77.7} & \textbf{40.2} & \textbf{43.2} & \textbf{23.4} & \textbf{18.5} & \textbf{51.1} \\
\midrule
\multicolumn{9}{c}{\textit{Qwen Series Models}} \\
\midrule
Qwen-2.5-0.5B~\cite{yang2024qwen2} & 36.7 & 15.8 & 4.8 & 2.8 & 0.0 & 0.3 & 12.5 \\
+ SimpleRL-Zoo & 49.5 & 34.4 & 10.3 & 8.9 & 0.0 & 0.7 & 22.5 \\
\rowcolor{colorful}
+ \textbf{R-TAP} & \textbf{54.3} & \textbf{39.8} & \textbf{16.6} & \textbf{14.2} & \textbf{6.4} & \textbf{6.9} & \textbf{28.3} \\
\cdashline{1-9}\noalign{\vskip 0.5ex}
Qwen-2.5-1.5B~\cite{yang2024qwen2} & 55.7 & 29.6 & 6.6 & 6.5 & 0.0 & 0.1 & 12.5 \\
+ SimpleRL-Zoo & 74.4 & 59.0 & 20.2 & 21.0 & 6.7 & 4.2 & 35.0 \\
\rowcolor{colorful}
+ \textbf{R-TAP} & \textbf{79.2} & \textbf{64.4} & \textbf{27.1} & \textbf{28.2} & \textbf{13.0} & \textbf{11.5} & \textbf{41.7} \\
\cdashline{1-9}\noalign{\vskip 0.5ex}
Qwen-2.5-7B~\cite{yang2024qwen2} & 88.2 & 64.6 & 25.7 & 30.1 & 3.3 & 0.3 & 30.0 \\
+ SimpleRL-Zoo & 91.7 & 78.2 & 38.6 & 40.4 & 20.0 & 15.6 & 62.5 \\
\rowcolor{colorful}
+ \textbf{R-TAP} & \textbf{96.0} & \textbf{83.5} & \textbf{44.8} & \textbf{46.9} & \textbf{28.1} & \textbf{22.3} & \textbf{69.4} \\
\cdashline{1-9}\noalign{\vskip 0.5ex}
Qwen-2.5-Math-7B~\cite{yang2024qwen2math} & 86.5 & 63.6 & 12.5 & 25.8 & 13.3 & 8.6 & 42.5 \\
+ SimpleRL-Zoo & 90.2 & 80.2 & 37.5 & 39.0 & 40.0 & 24.0 & 70.0 \\
\rowcolor{colorful}
+ \textbf{R-TAP} & \textbf{94.2} & \textbf{86.7} & \textbf{42.3} & \textbf{46.1} & \textbf{45.6} & \textbf{29.3} & \textbf{78.1} \\
\cdashline{1-9}\noalign{\vskip 0.5ex}
Qwen-2.5-14B~\cite{yang2024qwen2} & 91.6 & 65.4 & 24.3 & 33.5 & 6.7 & 3.4 & 37.5 \\
+ SimpleRL-Zoo & 94.4 & 80.2 & 40.4 & 44.9 & 23.3 & 14.2 & 57.6 \\
\rowcolor{colorful}
+ \textbf{R-TAP} & \textbf{97.8} & \textbf{86.3} & \textbf{48.1} & \textbf{52.3} & \textbf{30.5} & \textbf{19.8} & \textbf{61.7}  \\
\cdashline{1-9}\noalign{\vskip 0.5ex}
Qwen-2.5-32B~\cite{yang2024qwen2} & 92.9 & 68.6 & 27.9 & 31.1 & 10.0 & 4.5 & 45.0\\
+ SimpleRL-Zoo & 95.9 & 82.4 & 42.6 & 46.4 & 36.7 & 27.2 & 67.5 \\
\rowcolor{colorful}
+ \textbf{R-TAP} & \textbf{97.9} & \textbf{87.9} & \textbf{47.3} & \textbf{52.3} & \textbf{42.9} & \textbf{32.0} & \textbf{71.8} \\
\bottomrule
\end{tabular}
}
\vspace{-3mm}
\end{table*}

In summary, R-TAP provides broadly applicable mechanism for any LLMs and VLMs to enable recursive, confidence-guided reasoning within existing Think–Answer architectures. Furthermore, since the Confidence Generator is used only during training, R-TAP preserves the inference-time efficiency of standard single-pass models while equipping them with the ability to internally assess uncertainty and selectively refine their own reasoning for more stable yet faster. This makes R-TAP a scalable and practical solution for improving the stability, reliability, and depth of reasoning in modern models—bridging the gap between rigid single-pass inference and fully self-corrective reasoning systems.
\section{Experiments}

\subsection{Implementation Details}
We conduct training and evaluation of R-TAP primarily on NVIDIA A100 80GB GPUs. To ensure fast text generation during training, we employ vLLM~\cite{kwon2023efficient}, which is built on PagedAttention. In the pre-training step of Confidence Generator, vLLM~\cite{kwon2023efficient} is used to generate $N=128$ responses for each question for both LLMs and VLMs. We train the Confidence Generator under DeepSpeed engine with ZeRO-3~\cite{rajbhandari2020zero} with AdamW optimizer~\cite{loshchilov2018decoupled}, applying a linearly decayed learning rate from 1e-5 to 1e-6 to pre-train the confidence generator. Next step is training LLMs/VLMs as well as the Confidence Generator, where we also leverage vLLM~\cite{kwon2023efficient} for online text generation and use DeepSpeed with ZeRO-3 to update both models with a fixed learning rate of 1e-6. During this phase, we use $T=4$ recursion depths, for which we generate $G=12$ response outputs. From this setting, we update target models for 12 GRPO~\cite{shao2024deepseekmath} iterations per each training iteration by using $\epsilon=0.2$ and $\beta=0.04$ in \cref{eq:grpo}. For diverse responses, we set the generation hyperparameters to temperature=1.0, top-p=0.95, top-k=50, and repetition penalty=1.05. In addition, we set $\tau$ to 0.55 and we find its optimal $\tau$ by doing greedy search from zero to one value by 0.05 points. For evaluation, we remove the confidence generator and use only the trained LLMs/VLMs. We keep the default generation hyperparameters of backbone LLMs/VLMs. 

\begin{table*}[t!]
\vspace{-5mm}
\centering
\begin{minipage}{0.49\textwidth}
\centering
\vspace{-3mm}
\caption{Effect of each R-TAP component on LLM}
\vspace{-3mm}
\label{tab:rtap_llm}
\resizebox{\textwidth}{!}{
\begin{tabular}{lcccccccccc}
\toprule
Method &
$\mathbb{C}_{\phi}$ &
$R_{\text{Increase}}$ &
$R_{\text{Final}}$ &
$R_{\text{Answer}}$ &
AIME25 & HMMT Feb25 & OmniMath & GPQA & LiveCodeBench & Avg\\
\midrule

Baseline & \xmark & \xmark & \xmark & \cmark &
78.0 & 53.6 & 81.9 & 69.3 & 65.9 & 69.7 \\

R-TAP & \cmark & \cmark & \xmark & \cmark &
80.1 & 56.2 & 84.0 & 71.0 & 70.1 & 72.3 \\

R-TAP & \cmark & \xmark & \cmark & \cmark &
81.5 & 57.8 & 85.1 & 73.2 & 72.0 & 73.9 \\

\rowcolor{colorful}
R-TAP & \cmark & \cmark & \cmark & \cmark &
\textbf{83.7} & \textbf{60.3} & \textbf{86.2} &
\textbf{76.7} & \textbf{72.1} & \textbf{75.8} \\

R-TAP & \cmark & \cmark & \cmark & \xmark &
61.2 & 44.7 & 59.9 & 53.0 & 51.8 & 56.1 \\
\bottomrule
\end{tabular}
}
\end{minipage}
\hfill
\begin{minipage}{0.49\textwidth}
\centering
\vspace{-3mm}
\caption{Effect of each R-TAP component on VLM}
\vspace{-3mm}
\label{tab:rtap_vlm}
\resizebox{\textwidth}{!}{
\begin{tabular}{lcccccccccc}
\toprule
Method &
$\mathbb{C}_{\phi}$ &
$R_{\text{Increase}}$ &
$R_{\text{Final}}$ &
$R_{\text{Answer}}$ &
MMMU & MathVista & OlympiadBench & MathVision & MMMU-Pro & Avg \\
\midrule

Baseline & \xmark & \xmark & \xmark & \cmark &
73.6 & 74.0 & 62.6 & 49.0 & 52.0 & 62.2 \\

R-TAP & \cmark & \cmark & \xmark & \cmark &
76.1 & 78.0 & 66.4 & 53.0 & 54.4 & 65.4 \\

R-TAP & \cmark & \xmark & \cmark & \cmark &
77.8 & 80.2 & 68.3 & 55.3 & 58.1 & 67.9 \\

\rowcolor{colorful}
R-TAP & \cmark & \cmark & \cmark & \cmark &
\textbf{78.2} & \textbf{82.3} & \textbf{69.4} &
\textbf{56.8} & \textbf{59.2} & \textbf{69.2} \\

R-TAP & \cmark & \cmark & \cmark & \xmark &
53.2 & 50.7 & 45.9 & 41.3 & 52.4 & 48.7 \\
\bottomrule
\end{tabular}

}
\end{minipage}
\end{table*}

\begin{table*}[t!]
\centering
\begin{minipage}{0.49\textwidth}
\centering
\vspace{-3mm}
\caption{Confidence Estimation Methods for LLM}
\vspace{-3mm}
\label{tab:conf_llm}
\resizebox{\textwidth}{!}{
\begin{tabular}{lcccccc}
\toprule
LLMs & AIME25 & HMMT Feb25 & OmniMath & GPQA & LCB & Avg \\
\midrule
Phi-4-reasoning-plus       & 78.0 & 53.6 & 81.9 & 69.3 & 65.9 & 69.7 \\
Phi-4-reasoning-plus (a)   & 81.5 & 57.0 & 85.7 & 73.0 & 69.3 & 73.3 \\
Phi-4-reasoning-plus (b)   & 83.0 & 55.8 & 86.5 & 73.5 & 68.7 & 73.5 \\
Phi-4-reasoning-plus (c)   & 82.5 & 57.0 & 87.2 & 72.8 & 70.5 & 74.0 \\
Phi-4-reasoning-plus (d)   & \textbf{84.2} & 57.3 & \textbf{87.5} & 75.1 & 70.4 & 74.9 \\
\rowcolor{colorful}
\textbf{Phi-4-reasoning-plus-R-TAP} 
& 83.7 & \textbf{60.3} & 86.2 & \textbf{76.7} & \textbf{72.1} & \textbf{75.8} \\
\bottomrule
\end{tabular}

}
\end{minipage}
\hfill
\begin{minipage}{0.49\textwidth}
\centering
\vspace{-3mm}
\caption{Confidence estimation methods for VLM}
\vspace{-3mm}
\label{tab:conf_vlm}
\resizebox{\textwidth}{!}{
\begin{tabular}{lcccccc}
\toprule
VLMs & MMMU & MathVista & OlympiadBench & MathVision & MMMU-Pro & Avg \\
\midrule
R1V2-38B             & 73.6 & 74.0 & 62.6 & 49.0 & 52.0 & 62.2 \\
R1V2-38B (a)         & 75.1 & 78.0 & 65.8 & 53.5 & 57.6 & 66.0 \\
R1V2-38B (b)         & 76.3 & 77.5 & 67.2 & 53.0 & 58.5 & 66.5 \\
R1V2-38B (c)         & 77.0 & 79.0 & 68.5 & 54.2 & 57.3 & 67.2 \\
R1V2-38B (d)         & \textbf{79.5} & 80.3 & \textbf{70.8} & \textbf{58.2} & 55.7 & 68.9 \\
\rowcolor{colorful}
\textbf{R1V2-38B-R-TAP} 
& 78.2 & \textbf{82.3} & 69.4 & 56.8 & \textbf{59.2} & \textbf{69.2} \\
\bottomrule
\end{tabular}

}
\end{minipage}
\end{table*}

\begin{table*}[t!]
\centering
\begin{minipage}{0.49\textwidth}
\centering
\vspace{-3mm}
\caption{Recursion depths for LLM}
\vspace{-3mm}
\label{tab:iter_llm}
\resizebox{\textwidth}{!}{
\begin{tabular}{lccccccc}
\toprule
LLMs & $T$ & AIME25 & HMMT Feb25 & OmniMath & GPQA & LiveCodeBench & Avg \\
\midrule
Phi-4-reasoning-plus & 1 & 78.0 & 53.6 & 81.9 & 69.3 & 65.9 & 69.7 \\
Phi-4-reasoning-plus & 2 & 78.7 & 54.2 & 82.4 & 70.4 & 67.3 & 71.0 \\
Phi-4-reasoning-plus & 3 & 79.2 & 55.0 & 82.8 & 71.0 & 67.9 & 71.2 \\
Phi-4-reasoning-plus & 4 & 80.3 & 56.4 & 83.6 & 72.5 & 69.1 & 72.5 \\
\rowcolor{colorful}
\textbf{Phi-4-reasoning-plus-R-TAP} & 2 & 80.9 & 57.1 & 84.1 & 73.3 & 68.6 & 72.6 \\
\rowcolor{colorful}
\textbf{Phi-4-reasoning-plus-R-TAP} & 3 & 82.3 & 58.9 & 85.0 & 75.1 & 70.3 & 74.9 \\
\rowcolor{colorful}
\textbf{Phi-4-reasoning-plus-R-TAP} & 4 & \textbf{83.7} & \textbf{60.3} & \textbf{86.2} & \textbf{76.7} & \textbf{72.1} & \textbf{75.8} \\
\bottomrule
\end{tabular}

}
\end{minipage}
\hfill
\begin{minipage}{0.49\textwidth}
\centering
\vspace{-3mm}
\caption{Recursion depths for VLM}
\vspace{-3mm}
\label{tab:iter_vlm}
\resizebox{\textwidth}{!}{
\begin{tabular}{lccccccc}
\toprule
VLMs & $T$ & MMMU & MathVista & OlympiadBench & MathVision & MMMU-Pro & Avg \\
\midrule
R1V2-38B & 1 & 73.6 & 74.0 & 62.6 & 49.0 & 52.0 & 62.2 \\
R1V2-38B & 2 & 74.3 & 75.1 & 63.3 & 50.4 & 53.1 & 63.0 \\
R1V2-38B & 3 & 74.8 & 75.7 & 64.0 & 51.1 & 53.9 & 63.4 \\
R1V2-38B & 4 & 75.6 & 76.6 & 64.9 & 52.2 & 54.6 & 64.0 \\
\rowcolor{colorful}
\textbf{R1V2-38B-R-TAP} & 2 & 76.1 & 77.0 & 65.3 & 52.7 & 55.0 & 64.1 \\
\rowcolor{colorful}
\textbf{R1V2-38B-R-TAP} & 3 & 77.5 & 79.4 & 67.2 & 54.9 & 57.8 & 67.4 \\
\rowcolor{colorful}
\textbf{R1V2-38B-R-TAP} & 4 & \textbf{78.2} & \textbf{82.3} & \textbf{69.4} & \textbf{56.8} & \textbf{59.2} & \textbf{69.2} \\
\bottomrule
\end{tabular}

}
\end{minipage}
\vspace{-3mm}
\end{table*}

\begin{figure*}[t!]
\vspace{0mm}
    \centering
    \includegraphics[width=0.8\textwidth]{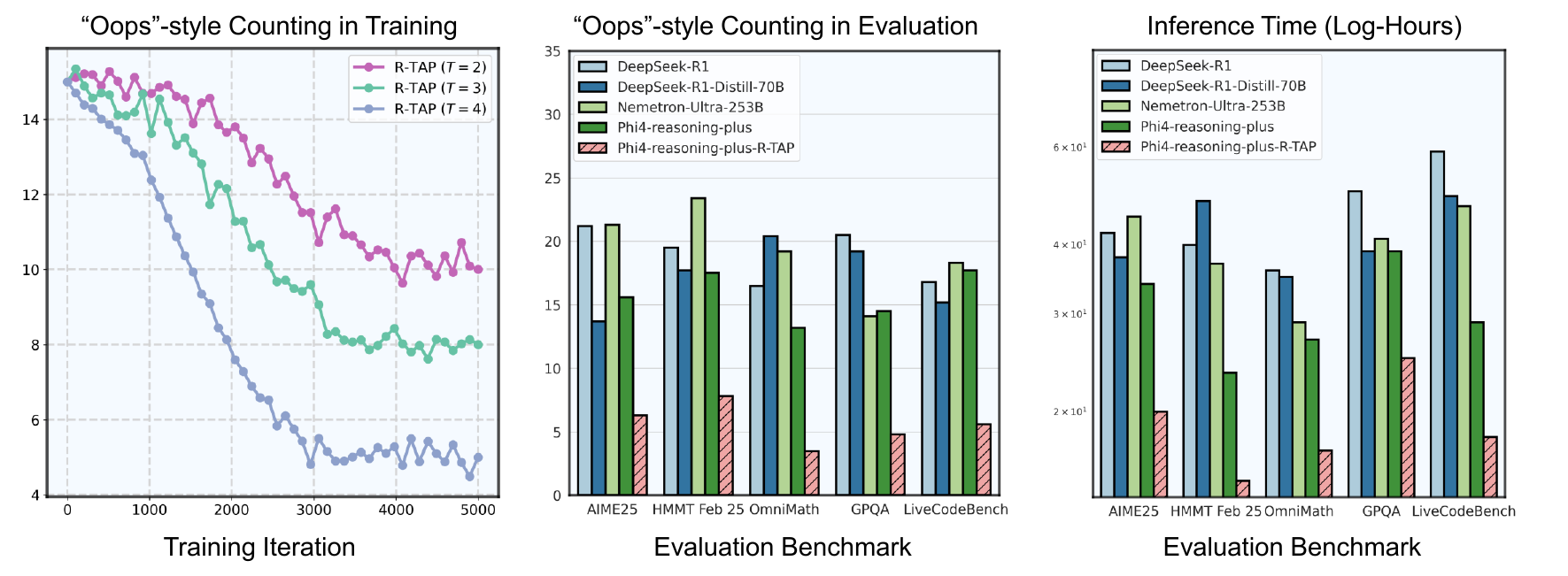}
    \vspace{-3mm}
    \caption{Impact of R-TAP on reducing the number of ``Oops''-style words -- which corresponds to the number of erroneous reasoning -- and its effect on substantially reducing inference time. (Left) Negative correlation between the number of erroneous reasoning and R-TAP train iterations. (Center) Evaluation result on the significant reduction of erroneous reasoning by applying R-TAP. (Right) Evaluation result on the substantial reduction of inference time due to the reduction of erroneous reasoning by applying R-TAP.}
    \label{fig:oops}
\vspace{-3mm}
\end{figure*}

\subsection{Validating R-TAP}
For LLMs, we train all the parameters, but for VLMs, we do not train vision encoder parts due to some observations of performance degradation. For selecting training dataset, we employ same training dataset on which LLMs/VLMs is trained for think-answer process, except some LLMs that do not release their own training dataset such as Phi-4-reasoning series~\cite{abdin2025phi}, MiMo~\cite{coreteam2025mimounlockingreasoningpotential}, Qwen3~\cite{yang2025qwen3technicalreport}, and AM-Thinking-v1~\cite{ji2025amthinkingv1advancingfrontierreasoning}. For these models, we instead gather Open-R1-Math (220K) and codeforce-cot (100K) in Open-R1~\cite{openr1}, and PRIME (481K)~\cite{cui2025process} covering math and code as well, and train them by using algorithm of R-TAP.

As shown in Fig.~\ref{fig:llm_bench}, R-TAP-applied Phi-4-reasoning models have shown dramatic improvements, thereby closing to OpenAI-o1 and -o3 models. Figure~\ref{fig:rew_curve} represents the reward graph of Phi-4-reasoning-plus~\cite{abdin2025phi} during training R-TAP, indicating stable training. To see more generalizability and applicability, we employ several LLMs and their dataset. First, we consider R1-Distill-Qwen-1.5B and -7B~\cite{guo2025deepseek} and Oat-Zero-1.5B, -7B~\cite{liu2025understanding}, AZR-Base-7B and AZR-Coder-7B~\cite{zhao2025absolute} as backbone models and we train them by R-TAP on Open-Reasoner-Zero-57K~\cite{hu2025open}. In addition, we apply R-TAP to SimpleRL-Zoo~\cite{zeng2025simplerl} in \cref{tab:simp} and PRIME~\cite{cui2025process} by using each their own curated data (57K and 481K, respectively). Remarkably,  \cref{tab:oat}-\cref{tab:prime} shows dramatic improvements of LLMs with R-TAP as well. We additionally employ recently released LLMs: Qwen3~\cite{yang2025qwen3technicalreport} and AM-Thinking-v1~\cite{ji2025amthinkingv1advancingfrontierreasoning}, and we observe their consistent dramatic performance improvements by R-TAP in \cref{tab:amthink}, compared with state-of-the-art open- and closed-source reasoning LLMs.

To extend its effectiveness to VLMs, we apply R-TAP to Skywork-R1V2~\cite{wei2025skywork} by Skywork-OR1-RL-120K, Geometry-3K, R1-OneVision-155K, and MMK12-16K. Notably, \cref{fig:vlm_bench} shows its consistent performance improvements despite multimodality. Besides, we employ Noisy-Rollout-7B~\cite{liu2025noisyrollout}, R1-OneVision-7B~\cite{yang2025r1}, and MM-Eureka-7B and -32B~\cite{meng2025mm} and train them on their own training dataset: Geometry-3K, R1-OneVision-155K, and MMK12-16K. \cref{tab:noisy}, \cref{tab:r1ov}, and \cref{tab:eureka} shows their performance improvements as well.

\subsection{Ablation Studies}
We conduct comprehensive ablation studies to validate the effect of each R-TAP component in \cref{tab:rtap_llm} and \cref{tab:rtap_vlm}, superiority of confidence estimation method in \cref{tab:conf_llm} and \cref{tab:conf_vlm}, and the effect of recursion depths in \cref{tab:iter_llm} and \cref{tab:iter_vlm}. In this study, we deal with Phi-4-reasoning-plus~\cite{abdin2025phi} for LLM and Skywork-R1V2~\cite{wei2025skywork} for VLM. We analyze the effect of removing confidence increase reward $R_{\text{Increase}}$, final answer confidence reward $R_{\text{Final}}$ and answer reward $R_{\text{Answer}}$. \cref{tab:rtap_llm} and \cref{tab:rtap_vlm} show that all components of R-TAP are essential for achieving highest accuracy in both language-only and multi-modal settings.

We further show the superiority of our Confidence Generator design by comparing with four recent confidence estimation approaches: (a) Calibration via ECE~\cite{guo2017calibration, geng2024survey} (b) Semantic Equivalence Entropy~\cite{kuhn2023semantic} (c) ``IDK'' token~\cite{cohen2024don} and (d) Iterative LLM generation~\cite{zhang2024calibrating, li2024think}. \cref{tab:conf_llm} and \cref{tab:conf_vlm} show that applying our method achieves highest accuracy. \cref{tab:iter_llm} and \cref{tab:iter_vlm} suggest that increasing recursion depths of Think-Answer from 1 to 4 times makes higher accuracy.

Finally, we additionally analyze how R-TAP affects the reduction of erroneous reasoning behaviors and inference efficiency, as summarized in \cref{fig:oops}. The left plot shows that the number of “Oops”-style tokens—our proxy for erroneous reasoning—monotonically decreases as training progresses, and deeper Think–Answer recursion (larger $T$) leads to faster and greater reduction. The center plot demonstrates that this trend generalizes to evaluation benchmarks: R-TAP consistently lowers erroneous reasoning across all datasets compared to baseline and the other state-of-the-art models. In addition, the right plot reveals that this reduction in reasoning errors directly translates to substantial improvements in inference efficiency, yielding significantly shorter inference time (log-hours) across all benchmarks. These results confirm that R-TAP not only improves accuracy but also makes reasoning more stable and computationally efficient. Besides, we describe detailed comparison between performance and computation complexity about decoding token count and training time in Appendix C.

\section{Discussion and Conclusion}
\label{sec:conclusion}
We introduced \textbf{R-TAP}, a method that augments LLMs and VLMs with a learned confidence generator and two rewards: recursive confidence increase and final-answer confidence reward. These components enable models to iteratively refine their think--answer trajectories, yielding substantial performance gains and narrowing the gap between smaller models and much larger ones. R-TAP supports up to $T$ recursive steps. However, due to the batch-dependent nature of current deep-learning implementations, all $T$ trajectories must be generated in parallel even when a confident answer emerges early. This simplifies parallelization but introduces significant computational and memory overhead. We hope that R-TAP encourages further research on confidence-aware iterative reasoning and supports the development of more efficient, trustworthy, and general-purpose reasoning systems for both LLMs and VLMs.

{
    \small
    \bibliographystyle{ieeenat_fullname}
    \bibliography{main}

@String(AAAI = {AAAI})

@misc{liu2024llavanext,
    title={LLaVA-NeXT: Improved reasoning, OCR, and world knowledge},
    url={https://llava-vl.github.io/blog/2024-01-30-llava-next/},
    author={Liu, Haotian and Li, Chunyuan and Li, Yuheng and Li, Bo and Zhang, Yuanhan and Shen, Sheng and Lee, Yong Jae},
    month={January},
    year={2024}
}

@article{yao2024minicpm,
  title={Minicpm-v: A gpt-4v level mllm on your phone},
  author={Yao, Yuan and Yu, Tianyu and Zhang, Ao and Wang, Chongyi and Cui, Junbo and Zhu, Hongji and Cai, Tianchi and Li, Haoyu and Zhao, Weilin and He, Zhihui and others},
  journal={arXiv preprint arXiv:2408.01800},
  year={2024}
}

@article{tong2024cambrian,
  title={Cambrian-1: A fully open, vision-centric exploration of multimodal llms},
  author={Tong, Shengbang and Brown, Ellis and Wu, Penghao and Woo, Sanghyun and Middepogu, Manoj and Akula, Sai Charitha and Yang, Jihan and Yang, Shusheng and Iyer, Adithya and Pan, Xichen and others},
  journal={arXiv preprint arXiv:2406.16860},
  year={2024}
}

@article{chen2024expanding,
  title={Expanding performance boundaries of open-source multimodal models with model, data, and test-time scaling},
  author={Chen, Zhe and Wang, Weiyun and Cao, Yue and Liu, Yangzhou and Gao, Zhangwei and Cui, Erfei and Zhu, Jinguo and Ye, Shenglong and Tian, Hao and Liu, Zhaoyang and others},
  journal={arXiv preprint arXiv:2412.05271},
  year={2024}
}

@article{young2024yi,
  title={Yi: Open foundation models by 01. ai},
  author={Young, Alex and Chen, Bei and Li, Chao and Huang, Chengen and Zhang, Ge and Zhang, Guanwei and Li, Heng and Zhu, Jiangcheng and Chen, Jianqun and Chang, Jing and others},
  journal={arXiv preprint arXiv:2403.04652},
  year={2024}
}

@misc{wang2024qwen2vl,
      title={Qwen2-VL: Enhancing Vision-Language Model's Perception of the World at Any Resolution}, 
      author={Peng Wang and Shuai Bai and Sinan Tan and Shijie Wang and Zhihao Fan and Jinze Bai and Keqin Chen and Xuejing Liu and Jialin Wang and Wenbin Ge and Yang Fan and Kai Dang and Mengfei Du and Xuancheng Ren and Rui Men and Dayiheng Liu and Chang Zhou and Jingren Zhou and Junyang Lin},
      year={2024},
      eprint={2409.12191},
      archivePrefix={arXiv},
      primaryClass={cs.CV},
      url={https://arxiv.org/abs/2409.12191}, 
}

@article{wang2024measuring,
  title={Measuring Multimodal Mathematical Reasoning with MATH-Vision Dataset},
  author={Wang, Ke and Pan, Junting and Shi, Weikang and Lu, Zimu and Zhan, Mingjie and Li, Hongsheng},
  journal={arXiv preprint arXiv:2402.14804},
  year={2024}
}

@article{team2023gemini,
  title={Gemini: a family of highly capable multimodal models},
  author={Team, Gemini and Anil, Rohan and Borgeaud, Sebastian and Wu, Yonghui and Alayrac, Jean-Baptiste and Yu, Jiahui and Soricut, Radu and Schalkwyk, Johan and Dai, Andrew M and Hauth, Anja and others},
  journal={arXiv preprint arXiv:2312.11805},
  year={2023}
}

@misc{claude3series2024,
  author = {{Anthropic}},
  title = {The Claude 3 Model Family: Opus, Sonnet, Haiku},
  year = {2024},
  howpublished = {\url{https://www.anthropic.com}},
  url = {https://www-cdn.anthropic.com/de8ba9b01c9ab7cbabf5c33b80b7bbc618857627/Model_Card_Claude_3.pdf}
}

@article{brown2020language,
  title={Language models are few-shot learners},
  author={Brown, Tom and Mann, Benjamin and Ryder, Nick and Subbiah, Melanie and Kaplan, Jared D and Dhariwal, Prafulla and Neelakantan, Arvind and Shyam, Pranav and Sastry, Girish and Askell, Amanda and others},
  journal={Advances in neural information processing systems},
  volume={33},
  pages={1877--1901},
  year={2020}
}

@article{touvron2023llama,
  title={Llama: Open and efficient foundation language models},
  author={Touvron, Hugo and Lavril, Thibaut and Izacard, Gautier and Martinet, Xavier and Lachaux, Marie-Anne and Lacroix, Timoth{\'e}e and Rozi{\`e}re, Baptiste and Goyal, Naman and Hambro, Eric and Azhar, Faisal and others},
  journal={arXiv preprint arXiv:2302.13971},
  year={2023}
}

@InProceedings{clip,
  title = 	 {Learning Transferable Visual Models From Natural Language Supervision},
  author =       {Radford, Alec and Kim, Jong Wook and Hallacy, Chris and Ramesh, Aditya and Goh, Gabriel and Agarwal, Sandhini and Sastry, Girish and Askell, Amanda and Mishkin, Pamela and Clark, Jack and Krueger, Gretchen and Sutskever, Ilya},
  booktitle = 	 {Proceedings of the 38th International Conference on Machine Learning},
  pages = 	 {8748--8763},
  year = 	 {2021},
  editor = 	 {Meila, Marina and Zhang, Tong},
  volume = 	 {139},
  series = 	 {Proceedings of Machine Learning Research},
  month = 	 {18--24 Jul},
  publisher =    {PMLR},
}

@inproceedings{
loshchilov2018decoupled,
title={Decoupled Weight Decay Regularization},
author={Ilya Loshchilov and Frank Hutter},
booktitle={International Conference on Learning Representations},
year={2019},
url={https://openreview.net/forum?id=Bkg6RiCqY7},
}

@inproceedings{lee2025multiverse,
  title={Multiverse: A multi-turn conversation benchmark for evaluating large vision and language models},
  author={Lee, Young-Jun and Lee, Byung-Kwan and Zhang, Jianshu and Hwang, Yechan and Ko, Byungsoo and Kim, Han-Gyu and Yao, Dongyu and Rong, Xuankun and Joo, Eojin and Han, Seung-Ho and others},
  booktitle={Proceedings of the IEEE/CVF International Conference on Computer Vision},
  pages={708--719},
  year={2025}
}

@article{lee2025refinebench,
  title={RefineBench: Evaluating Refinement Capability of Language Models via Checklists},
  author={Lee, Young-Jun and Kim, Seungone and Lee, Byung-Kwan and Moon, Minkyeong and Hwang, Yechan and Kim, Jong Myoung and Neubig, Graham and Welleck, Sean and Choi, Ho-Jin},
  journal={arXiv preprint arXiv:2511.22173},
  year={2025}
}

@article{lee2025genrecal,
  title={Genrecal: Generation after recalibration from large to small vision-language models},
  author={Lee, Byung-Kwan and Hachiuma, Ryo and Ro, Yong Man and Wang, Yu-Chiang Frank and Wu, Yueh-Hua},
  journal={arXiv preprint arXiv:2506.15681},
  year={2025}
}

@article{lee2025unified,
  title={Unified reinforcement and imitation learning for vision-language models},
  author={Lee, Byung-Kwan and Hachiuma, Ryo and Ro, Yong Man and Wang, Yu-Chiang Frank and Wu, Yueh-Hua},
  journal={arXiv preprint arXiv:2510.19307},
  year={2025}
}

@article{lee2025masking,
  title={Masking Teacher and Reinforcing Student for Distilling Vision-Language Models},
  author={Lee, Byung-Kwan and Wang, Yu-Chiang Frank and Hachiuma, Ryo},
  journal={arXiv preprint arXiv:2512.22238},
  year={2025}
}

@article{achiam2023gpt,
  title={GPT-4 technical report},
  author={Achiam, Josh and Adler, Steven and Agarwal, Sandhini and Ahmad, Lama and Akkaya, Ilge and Aleman, Florencia Leoni and Almeida, Diogo and Altenschmidt, Janko and Altman, Sam and Anadkat, Shyamal and others},
  journal={arXiv preprint arXiv:2303.08774},
  year={2023}
}

@article{lu2023mathvista,
  title={Mathvista: Evaluating mathematical reasoning of foundation models in visual contexts},
  author={Lu, Pan and Bansal, Hritik and Xia, Tony and Liu, Jiacheng and Li, Chunyuan and Hajishirzi, Hannaneh and Cheng, Hao and Chang, Kai-Wei and Galley, Michel and Gao, Jianfeng},
  journal={arXiv preprint arXiv:2310.02255},
  year={2023}
}

@article{liu2023hallusionbench,
  title={Hallusionbench: You see what you think? or you think what you see? an image-context reasoning benchmark challenging for gpt-4v (ision), llava-1.5, and other multi-modality models},
  author={Liu, Fuxiao and Guan, Tianrui and Li, Zongxia and Chen, Lichang and Yacoob, Yaser and Manocha, Dinesh and Zhou, Tianyi},
  journal={arXiv preprint arXiv:2310.14566},
  year={2023}
}

@article{yue2023mmmu,
  title={Mmmu: A massive multi-discipline multimodal understanding and reasoning benchmark for expert agi},
  author={Yue, Xiang and Ni, Yuansheng and Zhang, Kai and Zheng, Tianyu and Liu, Ruoqi and Zhang, Ge and Stevens, Samuel and Jiang, Dongfu and Ren, Weiming and Sun, Yuxuan and others},
  journal={arXiv preprint arXiv:2311.16502},
  year={2023}
}

@article{zhang2024mathverse,
  title={MathVerse: Does Your Multi-modal LLM Truly See the Diagrams in Visual Math Problems?},
  author={Zhang, Renrui and Jiang, Dongzhi and Zhang, Yichi and Lin, Haokun and Guo, Ziyu and Qiu, Pengshuo and Zhou, Aojun and Lu, Pan and Chang, Kai-Wei and Gao, Peng and others},
  journal={arXiv preprint arXiv:2403.14624},
  year={2024}
}

@article{dubey2024llama,
  title={The llama 3 herd of models},
  author={Dubey, Abhimanyu and Jauhri, Abhinav and Pandey, Abhinav and Kadian, Abhishek and Al-Dahle, Ahmad and Letman, Aiesha and Mathur, Akhil and Schelten, Alan and Yang, Amy and Fan, Angela and others},
  journal={arXiv preprint arXiv:2407.21783},
  year={2024}
}

@article{ouyang2022training,
  title={Training language models to follow instructions with human feedback},
  author={Ouyang, Long and Wu, Jeffrey and Jiang, Xu and Almeida, Diogo and Wainwright, Carroll and Mishkin, Pamela and Zhang, Chong and Agarwal, Sandhini and Slama, Katarina and Ray, Alex and others},
  journal={Advances in Neural Information Processing Systems},
  volume={35},
  pages={27730--27744},
  year={2022}
}

@article{li2024mini,
  title={Mini-Gemini: Mining the Potential of Multi-modality Vision Language Models},
  author={Li, Yanwei and Zhang, Yuechen and Wang, Chengyao and Zhong, Zhisheng and Chen, Yixin and Chu, Ruihang and Liu, Shaoteng and Jia, Jiaya},
  journal={arXiv preprint arXiv:2403.18814},
  year={2024}
}

@article{lee2024vlsi,
  title={VLsI: Verbalized Layers-to-Interactions from Large to Small Vision Language Models},
  author={Lee, Byung-Kwan and Hachiuma, Ryo and Wang, Yu-Chiang Frank and Ro, Yong Man and Wu, Yueh-Hua},
  journal={arXiv preprint arXiv:2412.01822},
  year={2024}
}

@article{lee2024phantom,
  title={Phantom of latent for large language and vision models},
  author={Lee, Byung-Kwan and Chung, Sangyun and Kim, Chae Won and Park, Beomchan and Ro, Yong Man},
  journal={arXiv preprint arXiv:2409.14713},
  year={2024}
}

@article{mckinzie2024mm1,
  title={Mm1: Methods, analysis \& insights from multimodal llm pre-training},
  author={McKinzie, Brandon and Gan, Zhe and Fauconnier, Jean-Philippe and Dodge, Sam and Zhang, Bowen and Dufter, Philipp and Shah, Dhruti and Du, Xianzhi and Peng, Futang and Weers, Floris and others},
  journal={arXiv preprint arXiv:2403.09611},
  year={2024}
}

@article{yue2024mmmu,
  title={Mmmu-pro: A more robust multi-discipline multimodal understanding benchmark},
  author={Yue, Xiang and Zheng, Tianyu and Ni, Yuansheng and Wang, Yubo and Zhang, Kai and Tong, Shengbang and Sun, Yuxuan and Yu, Botao and Zhang, Ge and Sun, Huan and others},
  journal={arXiv preprint arXiv:2409.02813},
  year={2024}
}

@article{kim2021distilling,
  title={Distilling robust and non-robust features in adversarial examples by information bottleneck},
  author={Kim, Junho and Lee, Byung-Kwan and Ro, Yong Man},
  journal={Advances in Neural Information Processing Systems},
  volume={34},
  pages={17148--17159},
  year={2021}
}

@inproceedings{lee2022masking,
  title={Masking adversarial damage: Finding adversarial saliency for robust and sparse network},
  author={Lee, Byung-Kwan and Kim, Junho and Ro, Yong Man},
  booktitle={Proceedings of the IEEE/CVF Conference on Computer Vision and Pattern Recognition},
  pages={15126--15136},
  year={2022}
}

@inproceedings{kim2023demystifying,
  title={Demystifying Causal Features on Adversarial Examples and Causal Inoculation for Robust Network by Adversarial Instrumental Variable Regression},
  author={Kim, Junho and Lee, Byung-Kwan and Ro, Yong Man},
  booktitle={Proceedings of the IEEE/CVF Conference on Computer Vision and Pattern Recognition},
  pages={12302--12312},
  year={2023}
}

@inproceedings{lee2023mitigating,
  title={Mitigating adversarial vulnerability through causal parameter estimation by adversarial double machine learning},
  author={Lee, Byung-Kwan and Kim, Junho and Ro, Yong Man},
  booktitle={Proceedings of the IEEE/CVF International Conference on Computer Vision},
  pages={4499--4509},
  year={2023}
}

@article{kim2023causal,
  title={Causal Unsupervised Semantic Segmentation},
  author={Kim, Junho and Lee, Byung-Kwan and Ro, Yong Man},
  journal={arXiv preprint arXiv:2310.07379},
  year={2023}
}

@misc{
lee2020towards,
title={Towards Adversarial Robustness of Bayesian Neural Network through Hierarchical Variational Inference},
author={Byung-Kwan Lee and Youngjoon Yu and Yong Man Ro},
year={2021},
url={https://openreview.net/forum?id=Cue2ZEBf12}
}

@article{lee2024collavo,
  title={CoLLaVO: Crayon Large Language and Vision mOdel},
  author={Lee, Byung-Kwan and Park, Beomchan and Kim, Chae Won and Ro, Yong Man},
  journal={arXiv preprint arXiv:2402.11248},
  year={2024}
}

@article{lee2024moai,
  title={MoAI: Mixture of All Intelligence for Large Language and Vision Models},
  author={Lee, Byung-Kwan and Park, Beomchan and Kim, Chae Won and Ro, Yong Man},
  journal={arXiv preprint arXiv:2403.07508},
  year={2024}
}

@inproceedings{rajbhandari2020zero,
  title={Zero: Memory optimizations toward training trillion parameter models},
  author={Rajbhandari, Samyam and Rasley, Jeff and Ruwase, Olatunji and He, Yuxiong},
  booktitle={SC20: International Conference for High Performance Computing, Networking, Storage and Analysis},
  pages={1--16},
  year={2020},
  organization={IEEE}
}

@article{meng2024simpo,
  title={Simpo: Simple preference optimization with a reference-free reward},
  author={Meng, Yu and Xia, Mengzhou and Chen, Danqi},
  journal={arXiv preprint arXiv:2405.14734},
  year={2024}
}

@article{austin2021program,
  title={Program synthesis with large language models},
  author={Austin, Jacob and Odena, Augustus and Nye, Maxwell and Bosma, Maarten and Michalewski, Henryk and Dohan, David and Jiang, Ellen and Cai, Carrie and Terry, Michael and Le, Quoc and others},
  journal={arXiv preprint arXiv:2108.07732},
  year={2021}
}

@article{wei2022chain,
  title={Chain-of-thought prompting elicits reasoning in large language models},
  author={Wei, Jason and Wang, Xuezhi and Schuurmans, Dale and Bosma, Maarten and Xia, Fei and Chi, Ed and Le, Quoc V and Zhou, Denny and others},
  journal={Advances in neural information processing systems},
  volume={35},
  pages={24824--24837},
  year={2022}
}

@article{ge2024convllava,
  title={ConvLLaVA: Hierarchical Backbones as Visual Encoder for Large Multimodal Models},
  author={Ge, Chunjiang and Cheng, Sijie and Wang, Ziming and Yuan, Jiale and Gao, Yuan and Song, Jun and Song, Shiji and Huang, Gao and Zheng, Bo},
  journal={arXiv preprint arXiv:2405.15738},
  year={2024}
}

@article{lee2024trol,
  title={TroL: Traversal of Layers for Large Language and Vision Models},
  author={Lee, Byung-Kwan and Chung, Sangyun and Kim, Chae Won and Park, Beomchan and Ro, Yong Man},
  journal={arXiv preprint arXiv:2406.12246},
  year={2024}
}

@article{chen2024evlm,
  title={EVLM: An Efficient Vision-Language Model for Visual Understanding},
  author={Chen, Kaibing and Shen, Dong and Zhong, Hanwen and Zhong, Huasong and Xia, Kui and Xu, Di and Yuan, Wei and Hu, Yifei and Wen, Bin and Zhang, Tianke and others},
  journal={arXiv preprint arXiv:2407.14177},
  year={2024}
}

@article{hu2024minicpm,
  title={Minicpm: Unveiling the potential of small language models with scalable training strategies},
  author={Hu, Shengding and Tu, Yuge and Han, Xu and He, Chaoqun and Cui, Ganqu and Long, Xiang and Zheng, Zhi and Fang, Yewei and Huang, Yuxiang and Zhao, Weilin and others},
  journal={arXiv preprint arXiv:2404.06395},
  year={2024}
}

@inproceedings{kwon2023efficient,
  title={Efficient memory management for large language model serving with pagedattention},
  author={Kwon, Woosuk and Li, Zhuohan and Zhuang, Siyuan and Sheng, Ying and Zheng, Lianmin and Yu, Cody Hao and Gonzalez, Joseph and Zhang, Hao and Stoica, Ion},
  booktitle={Proceedings of the 29th Symposium on Operating Systems Principles},
  pages={611--626},
  year={2023}
}

@article{yang2024qwen2,
  title={Qwen2. 5 technical report},
  author={Yang, An and Yang, Baosong and Zhang, Beichen and Hui, Binyuan and Zheng, Bo and Yu, Bowen and Li, Chengyuan and Liu, Dayiheng and Huang, Fei and Wei, Haoran and others},
  journal={arXiv preprint arXiv:2412.15115},
  year={2024}
}

@article{bai2025qwen2,
  title={Qwen2. 5-vl technical report},
  author={Bai, Shuai and Chen, Keqin and Liu, Xuejing and Wang, Jialin and Ge, Wenbin and Song, Sibo and Dang, Kai and Wang, Peng and Wang, Shijie and Tang, Jun and others},
  journal={arXiv preprint arXiv:2502.13923},
  year={2025}
}

@article{guo2025deepseek,
  title={Deepseek-r1: Incentivizing reasoning capability in llms via reinforcement learning},
  author={Guo, Daya and Yang, Dejian and Zhang, Haowei and Song, Junxiao and Zhang, Ruoyu and Xu, Runxin and Zhu, Qihao and Ma, Shirong and Wang, Peiyi and Bi, Xiao and others},
  journal={arXiv preprint arXiv:2501.12948},
  year={2025}
}

@article{shao2024deepseekmath,
  title={Deepseekmath: Pushing the limits of mathematical reasoning in open language models},
  author={Shao, Zhihong and Wang, Peiyi and Zhu, Qihao and Xu, Runxin and Song, Junxiao and Bi, Xiao and Zhang, Haowei and Zhang, Mingchuan and Li, YK and Wu, Y and others},
  journal={arXiv preprint arXiv:2402.03300},
  year={2024}
}

@article{chen2025sft,
  title={SFT or RL? An Early Investigation into Training R1-Like Reasoning Large Vision-Language Models},
  author={Chen, Hardy and Tu, Haoqin and Wang, Fali and Liu, Hui and Tang, Xianfeng and Du, Xinya and Zhou, Yuyin and Xie, Cihang},
  journal={arXiv preprint arXiv:2504.11468},
  year={2025}
}

@article{liu2025understanding,
  title={Understanding r1-zero-like training: A critical perspective},
  author={Liu, Zichen and Chen, Changyu and Li, Wenjun and Qi, Penghui and Pang, Tianyu and Du, Chao and Lee, Wee Sun and Lin, Min},
  journal={arXiv preprint arXiv:2503.20783},
  year={2025}
}

@article{huang2025vision,
  title={Vision-r1: Incentivizing reasoning capability in multimodal large language models},
  author={Huang, Wenxuan and Jia, Bohan and Zhai, Zijie and Cao, Shaosheng and Ye, Zheyu and Zhao, Fei and Xu, Zhe and Hu, Yao and Lin, Shaohui},
  journal={arXiv preprint arXiv:2503.06749},
  year={2025}
}

@article{peng2025lmm,
  title={Lmm-r1: Empowering 3b lmms with strong reasoning abilities through two-stage rule-based rl},
  author={Peng, Yingzhe and Zhang, Gongrui and Zhang, Miaosen and You, Zhiyuan and Liu, Jie and Zhu, Qipeng and Yang, Kai and Xu, Xingzhong and Geng, Xin and Yang, Xu},
  journal={arXiv preprint arXiv:2503.07536},
  year={2025}
}

@article{madaan2023self,
  title={Self-refine: Iterative refinement with self-feedback},
  author={Madaan, Aman and Tandon, Niket and Gupta, Prakhar and Hallinan, Skyler and Gao, Luyu and Wiegreffe, Sarah and Alon, Uri and Dziri, Nouha and Prabhumoye, Shrimai and Yang, Yiming and others},
  journal={Advances in neural information processing systems},
  volume={36},
  pages={46534--46594},
  year={2023}
}

@article{liu2025trust,
  title={Trust, But Verify: A Self-Verification Approach to Reinforcement Learning with Verifiable Rewards},
  author={Liu, Xiaoyuan and Liang, Tian and He, Zhiwei and Xu, Jiahao and Wang, Wenxuan and He, Pinjia and Tu, Zhaopeng and Mi, Haitao and Yu, Dong},
  journal={arXiv preprint arXiv:2505.13445},
  year={2025}
}

@article{liu2025noisyrollout,
  title={NoisyRollout: Reinforcing Visual Reasoning with Data Augmentation},
  author={Liu, Xiangyan and Ni, Jinjie and Wu, Zijian and Du, Chao and Dou, Longxu and Wang, Haonan and Pang, Tianyu and Shieh, Michael Qizhe},
  journal={arXiv preprint arXiv:2504.13055},
  year={2025}
}

@article{meng2025mm,
  title={MM-Eureka: Exploring the Frontiers of Multimodal Reasoning with Rule-based Reinforcement Learning},
  author={Meng, Fanqing and Du, Lingxiao and Liu, Zongkai and Zhou, Zhixiang and Lu, Quanfeng and Fu, Daocheng and Han, Tiancheng and Shi, Botian and Wang, Wenhai and He, Junjun and others},
  journal={arXiv preprint arXiv:2503.07365},
  year={2025}
}

@article{zhou2025r1,
  title={R1-Zero's" Aha Moment" in Visual Reasoning on a 2B Non-SFT Model},
  author={Zhou, Hengguang and Li, Xirui and Wang, Ruochen and Cheng, Minhao and Zhou, Tianyi and Hsieh, Cho-Jui},
  journal={arXiv preprint arXiv:2503.05132},
  year={2025}
}

@article{abdin2025phi,
  title={Phi-4-reasoning Technical Report},
  author={Abdin, Marah and Agarwal, Sahaj and Awadallah, Ahmed and Balachandran, Vidhisha and Behl, Harkirat and Chen, Lingjiao and de Rosa, Gustavo and Gunasekar, Suriya and Javaheripi, Mojan and Joshi, Neel and others},
  journal={arXiv preprint arXiv:2504.21318},
  year={2025}
}

@misc{openr1,
    title = {Open R1: A fully open reproduction of DeepSeek-R1},
    url = {https://github.com/huggingface/open-r1},
    author = {Hugging Face},
    month = {January},
    year = {2025}
}

@article{cui2025process,
  title={Process reinforcement through implicit rewards},
  author={Cui, Ganqu and Yuan, Lifan and Wang, Zefan and Wang, Hanbin and Li, Wendi and He, Bingxiang and Fan, Yuchen and Yu, Tianyu and Xu, Qixin and Chen, Weize and others},
  journal={arXiv preprint arXiv:2502.01456},
  year={2025}
}

@article{hu2025open,
  title={Open-reasoner-zero: An open source approach to scaling up reinforcement learning on the base model},
  author={Hu, Jingcheng and Zhang, Yinmin and Han, Qi and Jiang, Daxin and Zhang, Xiangyu and Shum, Heung-Yeung},
  journal={arXiv preprint arXiv:2503.24290},
  year={2025}
}

@article{zeng2025simplerl,
  title={Simplerl-zoo: Investigating and taming zero reinforcement learning for open base models in the wild},
  author={Zeng, Weihao and Huang, Yuzhen and Liu, Qian and Liu, Wei and He, Keqing and Ma, Zejun and He, Junxian},
  journal={arXiv preprint arXiv:2503.18892},
  year={2025}
}

@article{zhao2025absolute,
  title={Absolute Zero: Reinforced Self-play Reasoning with Zero Data},
  author={Zhao, Andrew and Wu, Yiran and Yue, Yang and Wu, Tong and Xu, Quentin and Lin, Matthieu and Wang, Shenzhi and Wu, Qingyun and Zheng, Zilong and Huang, Gao},
  journal={arXiv preprint arXiv:2505.03335},
  year={2025}
}

@misc{coreteam2025mimounlockingreasoningpotential,
      title={MiMo: Unlocking the Reasoning Potential of Language Model -- From Pretraining to Posttraining}, 
      author={Core Team and Bingquan Xia and Bowen Shen and Cici and Dawei Zhu and Di Zhang and Gang Wang and Hailin Zhang and Huaqiu Liu and Jiebao Xiao and Jinhao Dong and Liang Zhao and Peidian Li and Peng Wang and Shihua Yu and Shimao Chen and Weikun Wang and Wenhan Ma and Xiangwei Deng and Yi Huang and Yifan Song and Zihan Jiang and Bowen Ye and Can Cai and Chenhong He and Dong Zhang and Duo Zhang and Guoan Wang and Hao Tian and Haochen Zhao and Heng Qu and Hongshen Xu and Jun Shi and Kainan Bao and QingKai Fang and Kang Zhou and Kangyang Zhou and Lei Li and Menghang Zhu and Nuo Chen and Qiantong Wang and Shaohui Liu and Shicheng Li and Shuhao Gu and Shuhuai Ren and Shuo Liu and Sirui Deng and Weiji Zhuang and Weiwei Lv and Wenyu Yang and Xin Zhang and Xing Yong and Xing Zhang and Xingchen Song and Xinzhe Xu and Xu Wang and Yihan Yan and Yu Tu and Yuanyuan Tian and Yudong Wang and Yue Yu and Zhenru Lin and Zhichao Song and Zihao Yue},
      year={2025},
      eprint={2505.07608},
      archivePrefix={arXiv},
      primaryClass={cs.CL},
      url={https://arxiv.org/abs/2505.07608}, 
}

@article{wei2025skywork,
  title={Skywork R1V2: Multimodal Hybrid Reinforcement Learning for Reasoning},
  author={Wei, Yichen and Peng, Yi and Wang, Xiaokun and Qiu, Weijie and Shen, Wei and Xie, Tianyidan and Pei, Jiangbo and Zhang, Jianhao and Hao, Yunzhuo and Song, Xuchen and others},
  journal={arXiv preprint arXiv:2504.16656},
  year={2025}
}

@article{yang2025r1,
  title={R1-onevision: Advancing generalized multimodal reasoning through cross-modal formalization},
  author={Yang, Yi and He, Xiaoxuan and Pan, Hongkun and Jiang, Xiyan and Deng, Yan and Yang, Xingtao and Lu, Haoyu and Yin, Dacheng and Rao, Fengyun and Zhu, Minfeng and others},
  journal={arXiv preprint arXiv:2503.10615},
  year={2025}
}

@article{deng2025openvlthinker,
  title={Openvlthinker: An early exploration to complex vision-language reasoning via iterative self-improvement},
  author={Deng, Yihe and Bansal, Hritik and Yin, Fan and Peng, Nanyun and Wang, Wei and Chang, Kai-Wei},
  journal={arXiv preprint arXiv:2503.17352},
  year={2025}
}

@misc{chen2025r1v,
  author       = {Chen, Liang and Li, Lei and Zhao, Haozhe and Song, Yifan and Vinci},
  title        = {R1-V: Reinforcing Super Generalization Ability in Vision-Language Models with Less Than \$3},
  howpublished = {\url{https://github.com/Deep-Agent/R1-V}},
  note         = {Accessed: 2025-02-02},
  year         = {2025}
}

@article{chen2022program,
  title={Program of thoughts prompting: Disentangling computation from reasoning for numerical reasoning tasks},
  author={Chen, Wenhu and Ma, Xueguang and Wang, Xinyi and Cohen, William W},
  journal={arXiv preprint arXiv:2211.12588},
  year={2022}
}

@article{yao2023tree,
  title={Tree of thoughts: Deliberate problem solving with large language models},
  author={Yao, Shunyu and Yu, Dian and Zhao, Jeffrey and Shafran, Izhak and Griffiths, Tom and Cao, Yuan and Narasimhan, Karthik},
  journal={Advances in neural information processing systems},
  volume={36},
  pages={11809--11822},
  year={2023}
}

@inproceedings{besta2024graph,
  title={Graph of thoughts: Solving elaborate problems with large language models},
  author={Besta, Maciej and Blach, Nils and Kubicek, Ales and Gerstenberger, Robert and Podstawski, Michal and Gianinazzi, Lukas and Gajda, Joanna and Lehmann, Tomasz and Niewiadomski, Hubert and Nyczyk, Piotr and others},
  booktitle={Proceedings of the AAAI Conference on Artificial Intelligence},
  volume={38},
  number={16},
  pages={17682--17690},
  year={2024}
}

@article{shinn2023reflexion,
  title={Reflexion: Language agents with verbal reinforcement learning},
  author={Shinn, Noah and Cassano, Federico and Gopinath, Ashwin and Narasimhan, Karthik and Yao, Shunyu},
  journal={Advances in Neural Information Processing Systems},
  volume={36},
  pages={8634--8652},
  year={2023}
}

@article{wang2022self,
  title={Self-consistency improves chain of thought reasoning in language models},
  author={Wang, Xuezhi and Wei, Jason and Schuurmans, Dale and Le, Quoc and Chi, Ed and Narang, Sharan and Chowdhery, Aakanksha and Zhou, Denny},
  journal={arXiv preprint arXiv:2203.11171},
  year={2022}
}

@article{jaech2024openai,
  title={Openai o1 system card},
  author={Jaech, Aaron and Kalai, Adam and Lerer, Adam and Richardson, Adam and El-Kishky, Ahmed and Low, Aiden and Helyar, Alec and Madry, Aleksander and Beutel, Alex and Carney, Alex and others},
  journal={arXiv preprint arXiv:2412.16720},
  year={2024}
}

@article{patel2024aime,
  title={AIME: AI System Optimization via Multiple LLM Evaluators},
  author={Patel, Bhrij and Chakraborty, Souradip and Suttle, Wesley A and Wang, Mengdi and Bedi, Amrit Singh and Manocha, Dinesh},
  journal={arXiv preprint arXiv:2410.03131},
  year={2024}
}

@article{he2024olympiadbench,
  title={Olympiadbench: A challenging benchmark for promoting agi with olympiad-level bilingual multimodal scientific problems},
  author={He, Chaoqun and Luo, Renjie and Bai, Yuzhuo and Hu, Shengding and Thai, Zhen Leng and Shen, Junhao and Hu, Jinyi and Han, Xu and Huang, Yujie and Zhang, Yuxiang and others},
  journal={arXiv preprint arXiv:2402.14008},
  year={2024}
}

@misc{math500,
  title = {MATH-500},
  howpublished = {\url{https://huggingface.co/datasets/HuggingFaceH4/MATH-500}},
  note = {Accessed: 2025-05-16}
}

@misc{amc23,
  title = {AMC-23},
  howpublished = {\url{https://huggingface.co/datasets/knoveleng/AMC-23}},
  note = {Accessed: 2025-05-16}
}

@misc{minervamath,
  title = {Minerva-Math},
  howpublished = {\url{https://huggingface.co/datasets/knoveleng/Minerva-Math}},
  note = {Accessed: 2025-05-16}
}

@article{yang2024qwen2math,
  title={Qwen2. 5-math technical report: Toward mathematical expert model via self-improvement},
  author={Yang, An and Zhang, Beichen and Hui, Binyuan and Gao, Bofei and Yu, Bowen and Li, Chengpeng and Liu, Dayiheng and Tu, Jianhong and Zhou, Jingren and Lin, Junyang and others},
  journal={arXiv preprint arXiv:2409.12122},
  year={2024}
}

@misc{liu2025there,
  title={There May Not be Aha Moment in R1-Zero-like Training — A Pilot Study},
  author={Liu, Zichen and Chen, Changyu and Li, Wenjun and Pang, Tianyu and Du, Chao and Lin, Min},
  year={2025},
  howpublished={\url{https://oatllm.notion.site/oat-zero}},
  note={Notion Blog},
}

@inproceedings{rein2024gpqa,
  title={Gpqa: A graduate-level google-proof q\&a benchmark},
  author={Rein, David and Hou, Betty Li and Stickland, Asa Cooper and Petty, Jackson and Pang, Richard Yuanzhe and Dirani, Julien and Michael, Julian and Bowman, Samuel R},
  booktitle={First Conference on Language Modeling},
  year={2024}
}

@article{du2025supergpqa,
  title={Supergpqa: Scaling llm evaluation across 285 graduate disciplines},
  author={Du, Xinrun and Yao, Yifan and Ma, Kaijing and Wang, Bingli and Zheng, Tianyu and Zhu, King and Liu, Minghao and Liang, Yiming and Jin, Xiaolong and Wei, Zhenlin and others},
  journal={arXiv preprint arXiv:2502.14739},
  year={2025}
}

@article{dua2019drop,
  title={DROP: A reading comprehension benchmark requiring discrete reasoning over paragraphs},
  author={Dua, Dheeru and Wang, Yizhong and Dasigi, Pradeep and Stanovsky, Gabriel and Singh, Sameer and Gardner, Matt},
  journal={arXiv preprint arXiv:1903.00161},
  year={2019}
}

@inproceedings{wang2024mmlu,
  title={Mmlu-pro: A more robust and challenging multi-task language understanding benchmark},
  author={Wang, Yubo and Ma, Xueguang and Zhang, Ge and Ni, Yuansheng and Chandra, Abhranil and Guo, Shiguang and Ren, Weiming and Arulraj, Aaran and He, Xuan and Jiang, Ziyan and others},
  booktitle={The Thirty-eight Conference on Neural Information Processing Systems Datasets and Benchmarks Track},
  year={2024}
}

@article{zhou2023instruction,
  title={Instruction-following evaluation for large language models},
  author={Zhou, Jeffrey and Lu, Tianjian and Mishra, Swaroop and Brahma, Siddhartha and Basu, Sujoy and Luan, Yi and Zhou, Denny and Hou, Le},
  journal={arXiv preprint arXiv:2311.07911},
  year={2023}
}

@article{jain2024livecodebench,
  title={Livecodebench: Holistic and contamination free evaluation of large language models for code},
  author={Jain, Naman and Han, King and Gu, Alex and Li, Wen-Ding and Yan, Fanjia and Zhang, Tianjun and Wang, Sida and Solar-Lezama, Armando and Sen, Koushik and Stoica, Ion},
  journal={arXiv preprint arXiv:2403.07974},
  year={2024}
}

@article{hurst2024gpt,
  title={Gpt-4o system card},
  author={Hurst, Aaron and Lerer, Adam and Goucher, Adam P and Perelman, Adam and Ramesh, Aditya and Clark, Aidan and Ostrow, AJ and Welihinda, Akila and Hayes, Alan and Radford, Alec and others},
  journal={arXiv preprint arXiv:2410.21276},
  year={2024}
}

@misc{qwq32b,
    title = {QwQ-32B: Embracing the Power of Reinforcement Learning},
    url = {https://qwenlm.github.io/blog/qwq-32b/},
    author = {Qwen Team},
    month = {March},
    year = {2025}
}

@article{chen2021codex,
  title={Evaluating Large Language Models Trained on Code},
  author={Mark Chen and Jerry Tworek and Heewoo Jun and Qiming Yuan and Henrique Ponde de Oliveira Pinto and Jared Kaplan and Harri Edwards and Yuri Burda and Nicholas Joseph and Greg Brockman and Alex Ray and Raul Puri and Gretchen Krueger and Michael Petrov and Heidy Khlaaf and Girish Sastry and Pamela Mishkin and Brooke Chan and Scott Gray and Nick Ryder and Mikhail Pavlov and Alethea Power and Lukasz Kaiser and Mohammad Bavarian and Clemens Winter and Philippe Tillet and Felipe Petroski Such and Dave Cummings and Matthias Plappert and Fotios Chantzis and Elizabeth Barnes and Ariel Herbert-Voss and William Hebgen Guss and Alex Nichol and Alex Paino and Nikolas Tezak and Jie Tang and Igor Babuschkin and Suchir Balaji and Shantanu Jain and William Saunders and Christopher Hesse and Andrew N. Carr and Jan Leike and Josh Achiam and Vedant Misra and Evan Morikawa and Alec Radford and Matthew Knight and Miles Brundage and Mira Murati and Katie Mayer and Peter Welinder and Bob McGrew and Dario Amodei and Sam McCandlish and Ilya Sutskever and Wojciech Zaremba},
  year={2021},
  eprint={2107.03374},
  archivePrefix={arXiv},
  primaryClass={cs.LG}
}

@article{zeng2025acecoder,
  title={ACECODER: Acing Coder RL via Automated Test-Case Synthesis},
  author={Zeng, Huaye and Jiang, Dongfu and Wang, Haozhe and Nie, Ping and Chen, Xiaotong and Chen, Wenhu},
  journal={arXiv preprint arXiv:2502.01718},
  year={2025}
}

@article{code-r1,
  title={Code-R1: Reproducing R1 for Code with Reliable Rewards},
  author={Liu, Jiawei and Zhang, Lingming},
  howpublished={\url{https://github.com/ganler/code-r1}},
  year={2025}
}

@misc{hu2025openreasonerzeroopensourceapproach,
      title={Open-Reasoner-Zero: An Open Source Approach to Scaling Up Reinforcement Learning on the Base Model}, 
      author={Jingcheng Hu and Yinmin Zhang and Qi Han and Daxin Jiang and Xiangyu Zhang and Heung-Yeung Shum},
      year={2025},
      eprint={2503.24290},
      archivePrefix={arXiv},
      primaryClass={cs.LG},
      url={https://arxiv.org/abs/2503.24290}, 
}

@article{grattafiori2024llama,
  title={The llama 3 herd of models},
  author={Grattafiori, Aaron and Dubey, Abhimanyu and Jauhri, Abhinav and Pandey, Abhinav and Kadian, Abhishek and Al-Dahle, Ahmad and Letman, Aiesha and Mathur, Akhil and Schelten, Alan and Vaughan, Alex and others},
  journal={arXiv preprint arXiv:2407.21783},
  year={2024}
}

@mastersthesis{jiang2024identifying,
  title={Identifying and mitigating vulnerabilities in llm-integrated applications},
  author={Jiang, Fengqing},
  year={2024},
  school={University of Washington}
}

@article{cobbe2021gsm8k,
  title={Training Verifiers to Solve Math Word Problems},
  author={Cobbe, Karl and Kosaraju, Vineet and Bavarian, Mohammad and Chen, Mark and Jun, Heewoo and Kaiser, Lukasz and Plappert, Matthias and Tworek, Jerry and Hilton, Jacob and Nakano, Reiichiro and Hesse, Christopher and Schulman, John},
  journal={arXiv preprint arXiv:2110.14168},
  year={2021}
}

@inproceedings{coignion2024performance,
  title={A performance study of llm-generated code on leetcode},
  author={Coignion, Tristan and Quinton, Cl{\'e}ment and Rouvoy, Romain},
  booktitle={Proceedings of the 28th International Conference on Evaluation and Assessment in Software Engineering},
  pages={79--89},
  year={2024}
}

@article{yuan2024advancing,
  title={Advancing llm reasoning generalists with preference trees},
  author={Yuan, Lifan and Cui, Ganqu and Wang, Hanbin and Ding, Ning and Wang, Xingyao and Deng, Jia and Shan, Boji and Chen, Huimin and Xie, Ruobing and Lin, Yankai and others},
  journal={arXiv preprint arXiv:2404.02078},
  year={2024}
}

@article{ahmadian2024back,
  title={Back to basics: Revisiting reinforce style optimization for learning from human feedback in llms},
  author={Ahmadian, Arash and Cremer, Chris and Gall{\'e}, Matthias and Fadaee, Marzieh and Kreutzer, Julia and Pietquin, Olivier and {\"U}st{\"u}n, Ahmet and Hooker, Sara},
  journal={arXiv preprint arXiv:2402.14740},
  year={2024}
}

@article{qiao2024we,
  title={We-math: Does your large multimodal model achieve human-like mathematical reasoning?},
  author={Qiao, Runqi and Tan, Qiuna and Dong, Guanting and Wu, Minhui and Sun, Chong and Song, Xiaoshuai and GongQue, Zhuoma and Lei, Shanglin and Wei, Zhe and Zhang, Miaoxuan and others},
  journal={arXiv preprint arXiv:2407.01284},
  year={2024}
}

@article{zhang2025r1,
  title={R1-vl: Learning to reason with multimodal large language models via step-wise group relative policy optimization},
  author={Zhang, Jingyi and Huang, Jiaxing and Yao, Huanjin and Liu, Shunyu and Zhang, Xikun and Lu, Shijian and Tao, Dacheng},
  journal={arXiv preprint arXiv:2503.12937},
  year={2025}
}

@misc{arenahard2024,
    title = {From Live Data to High-Quality Benchmarks: The Arena-Hard Pipeline},
    url = {https://lmsys.org/blog/2024-04-19-arena-hard/},
    author = {Tianle Li*, et al.},
    month = {April},
    year = {2024}
}

@misc{ji2025amthinkingv1advancingfrontierreasoning,
      title={AM-Thinking-v1: Advancing the Frontier of Reasoning at 32B Scale}, 
      author={Yunjie Ji and Xiaoyu Tian and Sitong Zhao and Haotian Wang and Shuaiting Chen and Yiping Peng and Han Zhao and Xiangang Li},
      year={2025},
      eprint={2505.08311},
      archivePrefix={arXiv},
      primaryClass={cs.CL},
      url={https://arxiv.org/abs/2505.08311}, 
}

@misc{yang2025qwen3technicalreport,
      title={Qwen3 Technical Report}, 
      author={An Yang and Anfeng Li, et al},
      year={2025},
      eprint={2505.09388},
      archivePrefix={arXiv},
      primaryClass={cs.CL},
      url={https://arxiv.org/abs/2505.09388}, 
}

@misc{gui2025adora,
  title={Training Reasoning Model with Dynamic Advantage Estimation on Reinforcement Learning},
  author={Lujun Gui and Qingnan Ren},
  year={2025},
  note={Notion Blog},
}

@article{wang2025sota,
  title={SoTA with Less: MCTS-Guided Sample Selection for Data-Efficient Visual Reasoning Self-Improvement},
  author={Wang, Xiyao and Yang, Zhengyuan and Feng, Chao and Lu, Hongjin and Li, Linjie and Lin, Chung-Ching and Lin, Kevin and Huang, Furong and Wang, Lijuan},
  journal={arXiv preprint arXiv:2504.07934},
  year={2025}
}

@misc{wu2024deepseekvl2mixtureofexpertsvisionlanguagemodels,
      title={DeepSeek-VL2: Mixture-of-Experts Vision-Language Models for Advanced Multimodal Understanding},
      author={Zhiyu Wu and Xiaokang Chen and Zizheng Pan and Xingchao Liu and Wen Liu and Damai Dai and Huazuo Gao and Yiyang Ma and Chengyue Wu and Bingxuan Wang and Zhenda Xie and Yu Wu and Kai Hu and Jiawei Wang and Yaofeng Sun and Yukun Li and Yishi Piao and Kang Guan and Aixin Liu and Xin Xie and Yuxiang You and Kai Dong and Xingkai Yu and Haowei Zhang and Liang Zhao and Yisong Wang and Chong Ruan},
      year={2024},
      eprint={2412.10302},
      archivePrefix={arXiv},
      primaryClass={cs.CV},
      url={https://arxiv.org/abs/2412.10302},
}

@misc{bercovich2025llamanemotronefficientreasoningmodels,
      title={Llama-Nemotron: Efficient Reasoning Models}, 
      author={Akhiad Bercovich and Itay Levy et al.},
      year={2025},
      eprint={2505.00949},
      archivePrefix={arXiv},
      primaryClass={cs.CL},
      url={https://arxiv.org/abs/2505.00949}, 
}

@misc{gpto3minisyscard,
    author= {{OpenAI}},
    year  = {2025},
    title = {OpenAI o3-mini System Card},
    note  = {\url{https://cdn.openai.com/o3-mini-system-card-feb10.pdf}, 
             Last accessed on 2025-05-16},
}

@inproceedings{guo2017calibration,
  title={On calibration of modern neural networks},
  author={Guo, Chuan and Pleiss, Geoff and Sun, Yu and Weinberger, Kilian Q},
  booktitle={International conference on machine learning},
  pages={1321--1330},
  year={2017},
  organization={PMLR}
}

@inproceedings{geng2024survey,
  title={A survey of confidence estimation and calibration in large language models},
  author={Geng, Jiahui and Cai, Fengyu and Wang, Yuxia and Koeppl, Heinz and Nakov, Preslav and Gurevych, Iryna},
  booktitle={Proceedings of the 2024 Conference of the North American Chapter of the Association for Computational Linguistics: Human Language Technologies (Volume 1: Long Papers)},
  pages={6577--6595},
  year={2024}
}

@article{kuhn2023semantic,
  title={Semantic uncertainty: Linguistic invariances for uncertainty estimation in natural language generation},
  author={Kuhn, Lorenz and Gal, Yarin and Farquhar, Sebastian},
  journal={arXiv preprint arXiv:2302.09664},
  year={2023}
}

@article{cohen2024don,
  title={I Don't Know: Explicit Modeling of Uncertainty with an [IDK] Token},
  author={Cohen, Roi and Dobler, Konstantin and Biran, Eden and de Melo, Gerard},
  journal={Advances in Neural Information Processing Systems},
  volume={37},
  pages={10935--10958},
  year={2024}
}

@article{zhang2024calibrating,
  title={Calibrating the confidence of large language models by eliciting fidelity},
  author={Zhang, Mozhi and Huang, Mianqiu and Shi, Rundong and Guo, Linsen and Peng, Chong and Yan, Peng and Zhou, Yaqian and Qiu, Xipeng},
  journal={arXiv preprint arXiv:2404.02655},
  year={2024}
}

@article{li2024think,
  title={Think twice before trusting: Self-detection for large language models through comprehensive answer reflection},
  author={Li, Moxin and Wang, Wenjie and Feng, Fuli and Zhu, Fengbin and Wang, Qifan and Chua, Tat-Seng},
  journal={arXiv preprint arXiv:2403.09972},
  year={2024}
}
}

\clearpage
\setcounter{page}{1}
\appendix
\onecolumn

\clearpage
\section{Full Responses for Multimodal Example}
\label{app:A}
\begin{figure*}[h!]
\vspace{0mm}
    \centering
    \includegraphics[width=\linewidth]{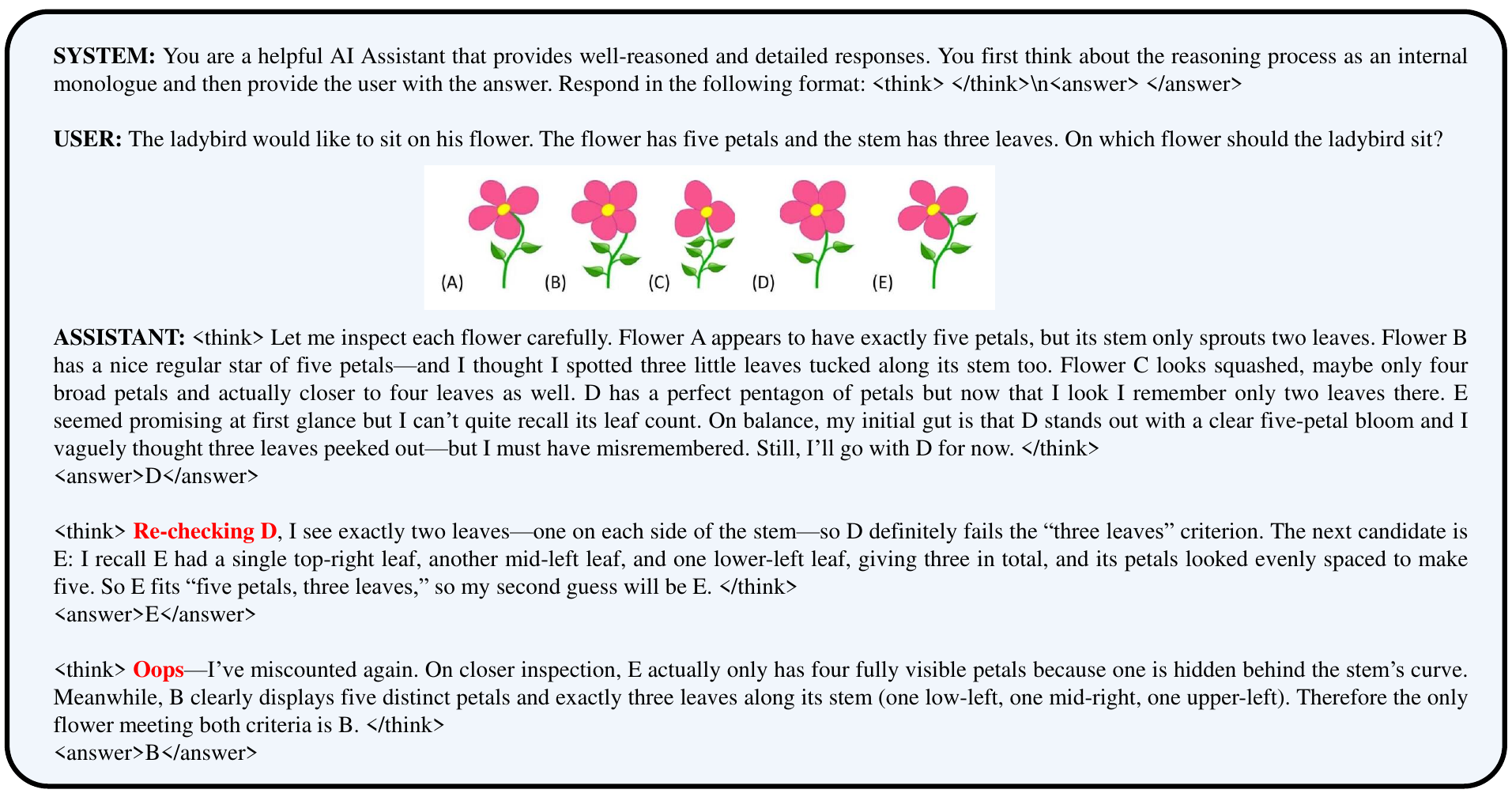}
    \caption{Qualitative multimodal example of recursive think–answer process. The model refines its answer over multiple reasoning cycles by correcting initial counting errors and verifying object-level visual details, ultimately identifying the only flower that meets both criteria.}
    \label{fig:qual_mm}
\vspace{0mm}
\end{figure*}


\section{Algorithm of Recursive Think-Answer Process}
\label{app:B}
\setlength{\textfloatsep}{0pt}
\begin{algorithm}[h!]
\caption{Recursive Think-Answer Process for LLMs and VLMs}
\label{alg:1}
\begin{algorithmic}[1]
\REQUIRE Pre-trained Confidence Generator $\mathcal{C}_{\phi}$ and Pre-trained LLMs/VLMs $\pi_{\theta_{\text{init}}}$
\STATE Set reference model $\pi_{\text{ref}} \gets \pi_{\theta_{\text{init}}}$
\STATE Set the training model $\pi_{\theta} \gets \pi_{\theta_{\text{init}}}$
\FOR{sample a batch $\mathcal{B}$ in Dataset}
    \STATE Copy and freeze model $\pi_{\theta_{\text{old}}} \gets \pi_\theta$
    \STATE Sample $G$ outputs $\{o_i\}_{i=1}^{G} \sim \pi_{\theta_{\text{old}}}(\cdot | q)$, until $M$ maximum recursive.
    \STATE Extract the responses $G'$ in $G$ outputs until the early correct responses.
    \STATE Replace the responses: $G \gets G'$
    \STATE Compute rewards and advantages for $G$ outputs by confidence generator \& answer parsing
    \FOR{Confidence generator Updating iteration $= 1, 2, \cdots, \mu$}
        \STATE Update $\mathcal{C}_{\phi}$ by using \cref{eq:conf_train}
    \ENDFOR
    \FOR{LLMs/VLMs Updating iteration $= 1, 2, \cdots, \mu$}
        \STATE Update $\pi_\theta$ by using Equation \cref{eq:grpo}
    \ENDFOR
\ENDFOR
\end{algorithmic}
\end{algorithm}

\clearpage
\section{Detailed Comparison between Performance and Computation Complexity}
\label{app:C}

\paragraph{Comparison with Related Refinement and Verification Methods.}

We compare R-TAP against four closely related baselines: (a) Reflexion~\cite{shinn2023reflexion}, (b) Self-Consistency~\cite{wang2022self}, (c) Self-Refine~\cite{madaan2023self}, and (d) Self-Verification~\cite{liu2025trust}. All methods are evaluated under the same output token budget. Unlike iterative refinement or verification-based approaches, which rely on explicit self-feedback loops, majority voting, or external verification during inference, R-TAP learns when to continue or terminate reasoning via reinforcement learning. As a result, it removes the need for repeated trial-and-error generation at deployment. Although the confidence generator in R-TAP is trained with binary supervision, it does not function as a hard 0/1 classifier at inference. Instead, it produces a continuous signal in $[0,1]$, which enables the model to measure confidence improvement across recursive steps and regulate reasoning depth smoothly.

\begin{table}[h!]
\vspace{-3mm}
\centering
\resizebox{\linewidth}{!}{
\renewcommand{\tabcolsep}{2mm}
\begin{tabular}{lcccccccccc}
\toprule
                     & MaxTokenLimit & AIME25 & HMMT Feb25 & OmniMath & GPQA & LCB  & Avg  & Oops-style count & Decoding token count & Training Time \\
\midrule
Phi-4-reasoning-plus & 32768           & 78.0   & 53.6       & 81.9     & 69.3 & 65.9 & 69.7 & 15.7             & 14509.7              & -                             \\
\midrule
w/ Reflexion~\cite{shinn2023reflexion}         & 32768           & 80.2   & 56.1       & 83.4     & 71.3 & 66.5 & 71.5 & 18.9             & 15230.4              & 51 hours                      \\
w/ Self-Consistency~\cite{wang2022self}  & 32768           & 81.1   & 55.8       & 84.0     & 72.2 & 66.4 & 71.9 & 16.8             & 16012.9              & 58 hours                      \\
w/ Self-Refine~\cite{madaan2023self}       & 32768           & 78.5   & 54.9       & 83.1     & 71.0 & 68.5 & 71.2 & 14.3             & 13890.6              & 45 hours                      \\
w/ Self-Verification~\cite{liu2025trust} & 32768           & 82.4   & 57.6       & 85.7     & 73.5 & 64.8 & 72.8 & 17.5             & 14890.3              & 51 hours                      \\
\rowcolor{colorful}
R-TAP                & 32768           & \textbf{83.7}   & \textbf{60.3}       &\textbf{ 86.2}     & \textbf{76.7} &\textbf{ 72.1 }& \textbf{75.8} & \textbf{5.6}              & \textbf{4378.8}               & \textbf{28 hours}    \\      
\bottomrule
\end{tabular}
}
\end{table}
\begin{table}[h!]
\centering
\vspace{-7mm}
\resizebox{\linewidth}{!}{
\renewcommand{\tabcolsep}{2mm}
\begin{tabular}{lcccccccccc}
\toprule
                     & MaxTokenLimit & MMMU & MathVista & OlympiadBench & MathVision & MMMU-Pro & Avg  & Oops-style count & Output token count & Training Time \\
\midrule
R1V2-38B             & 12000         & 73.6 & 74.0      & 62.6          & 49.0       & 52.0     & 62.2 & 17.2             & 9890.7               & -                             \\
\midrule
w/ Reflexion~\cite{shinn2023reflexion}         & 12000         & 74.5 & 74.8      & 63.0          & 49.5       & 53.7     & 63.1 & 17.5             & 10989.3              & 57 hours                      \\
w/ Self-Consistency~\cite{wang2022self}  & 12000         & 75.0 & 76.0      & 64.0          & 50.0       & 52.5     & 63.5 & 18.6             & 10678.5              & 60 hours                      \\
w/ Self-Refine~\cite{madaan2023self}       & 12000         & 73.8 & 74.2      & 62.0          & 49.0       & 55.5     & 62.9 & 15.3             & 9938.2               & 59 hours                      \\
w/ Self-Verification~\cite{liu2025trust} & 12000         & 76.5 & 77.0      & 65.0          & 51.0       & 53.0     & 64.5 & 17.8             & 10183.8              & 58 hours                      \\
\rowcolor{colorful}
R-TAP                & 12000         & \textbf{78.2} & \textbf{82.3}      & \textbf{69.4}          & \textbf{56.8}       & \textbf{59.2}     & \textbf{69.2} & \textbf{8.5}              & \textbf{5789.4 }              & \textbf{39 hours}                       \\
\bottomrule
\end{tabular}
}
\vspace{-5mm}
\end{table}

\paragraph{Token Efficiency and Computational Cost.}

R-TAP does not increase total output tokens. On the contrary, it substantially reduces them. While recursive sampling is introduced during training, the learned policy discourages unnecessary self-corrections at inference time. As shown in this table, R-TAP reduces output tokens by approximately 2--3$\times$ compared to self-consistency, self-refine, and verification-based baselines under the same output token budget, while achieving higher accuracy.

\begin{table}[h!]
\vspace{-3mm}
\centering
\resizebox{\linewidth}{!}{
\renewcommand{\tabcolsep}{5mm}
\begin{tabular}{cccccccccc}
\toprule
T & G  & AIME25 & HMMT Feb25 & OmniMath & GPQA & LiveCodeBench & Avg  & Output token count & Training Time \\
\midrule
1 & 4  & 76.1   & 52.1       & 80.4     & 67.9 & 64.4          & 68.2 & 15034.9              & 51 hours                      \\
1 & 8  & 77.2   & 52.9       & 81.2     & 68.6 & 65.2          & 69.0 & 15210.7              & 52 hours                      \\
1 & 12 & 78.0   & 53.6       & 81.9     & 69.3 & 65.9          & 69.7 & 14872.5              & 58 hours                      \\
2 & 12 & 80.9   & 57.1       & 84.1     & 73.3 & 68.6          & 72.6 & 9258.9               & 42 hours                      \\
3 & 12 & 82.3   & 58.9       & 85.0     & 75.1 & 70.3          & 74.9 & 6127.4               & 34 hours                      \\
\rowcolor{colorful}
4 & 12 & 83.7   & 60.3       & 86.2     & 76.7 & 72.1          & 75.8 & 4378.8               & 28 hours                      \\
4 & 8  & 83.0   & 59.7       & 85.7     & 76.0 & 71.4          & 75.2 & 4442.3               & 25 hours                      \\
4 & 4  & 82.2   & 58.9       & 85.0     & 75.2 & 70.6          & 74.4 & 4321.6               & 22 hours                      \\      
\bottomrule
\end{tabular}
}
\end{table}
\begin{table}[h!]
\centering
\vspace{-7mm}
\resizebox{\linewidth}{!}{
\renewcommand{\tabcolsep}{5mm}
\begin{tabular}{cccccccccc}
\toprule
T & G  & MMMU & MathVista & OlympiadBench & MathVision & MMMU-Pro & Avg  & Decoding token count & Training Time \\
\midrule
1 & 4  & 72.4 & 72.6      & 61.2          & 47.8       & 50.6     & 60.9 & 10342.7              & 55 hours                      \\
1 & 8  & 73.1 & 73.4      & 62.0          & 48.5       & 51.4     & 61.7 & 9927.3               & 62 hours                      \\
1 & 12 & 73.6 & 74.0      & 62.6          & 49.0       & 52.0     & 62.2 & 10168.9              & 68 hours                      \\
2 & 12 & 76.1 & 77.0      & 65.3          & 52.7       & 55.0     & 64.1 & 8234.6               & 55 hours                      \\
3 & 12 & 77.5 & 79.4      & 67.2          & 54.9       & 57.8     & 67.4 & 6912.1               & 46 hours                      \\
\rowcolor{colorful}
4 & 12 & 78.2 & 82.3      & 69.4          & 56.8       & 59.2     & 69.2 & 5789.4               & 39 hours                      \\
4 & 8  & 77.8 & 81.7      & 68.8          & 56.4       & 58.7     & 68.7 & 5698.2               & 34 hours                      \\
4 & 4  & 77.1 & 80.8      & 67.9          & 55.6       & 57.8     & 67.8 & 5861.5               & 30 hours                     \\
\bottomrule
\end{tabular}
}
\vspace{-5mm}
\end{table}

\paragraph{Effect of Majority Voting.}

We further evaluate self-consistency with varying voting numbers ($N$), under identical output token budgets (32768 for Phi-4-reasoning-plus and 12000 for R1V2-38B). As expected, increasing $N$ consistently improves performance across all models. However, R-TAP-trained models already achieve strong performance with $N=1$, and additional voting yields only marginal gains. This indicates that R-TAP does not replace self-consistency; rather, it learns a more stable single-sample reasoning policy, thereby reducing reliance on majority voting during inference.

\begin{table}[h!]
\vspace{-3mm}
\centering
\begin{minipage}[t]{0.49\linewidth}
\centering
\resizebox{\linewidth}{!}{
\renewcommand{\tabcolsep}{5mm}
\begin{tabular}{ccccccc}
\toprule
Voting N & AIME25 & HMMT Feb25 & OmniMath & GPQA & LiveCodeBench & Avg  \\
\midrule
1        & 83.7   & 60.3       & 86.2     & 76.7 & 72.1          & 75.8 \\
3        & 83.9   & 60.5       & 86.4     & 76.9 & 72.3          & 76.0 \\
5        & 84.0   & 60.6       & 86.5     & 77.0 & 72.4          & 76.1 \\
7        & 84.1   & 60.7       & 86.6     & 77.1 & 72.5          & 76.2 \\
9        & 84.1   & 60.7       & 86.6     & 77.1 & 72.5          & 76.2 \\
12       & 84.2   & 60.8       & 86.7     & 77.2 & 72.6          & 76.3 \\
15       & 84.2   & 60.8       & 86.7     & 77.2 & 72.6          & 76.3 \\
18       & 84.2   & 60.8       & 86.7     & 77.2 & 72.6          & 76.3 \\
\bottomrule
\end{tabular}
}
\end{minipage}
\hfill
\begin{minipage}[t]{0.49\linewidth}
\centering
\resizebox{\linewidth}{!}{
\renewcommand{\tabcolsep}{5mm}
\begin{tabular}{ccccccc}
\toprule
Voting N & MMMU & MathVista & OlympiadBench & MathVision & MMMU-Pro & Avg  \\
\midrule
1        & 78.2 & 82.3      & 69.4          & 56.8       & 59.2     & 69.2 \\
3        & 78.3 & 82.4      & 69.5          & 56.9       & 59.3     & 69.3 \\
5        & 78.3 & 82.4      & 69.5          & 56.9       & 59.3     & 69.3 \\
7        & 78.4 & 82.5      & 69.6          & 57         & 59.4     & 69.4 \\
9        & 78.5 & 82.6      & 69.7          & 57.1       & 59.5     & 69.5 \\
12       & 78.5 & 82.6      & 69.7          & 57.1       & 59.5     & 69.5 \\
15       & 78.5 & 82.6      & 69.7          & 57.1       & 59.5     & 69.5 \\
18       & 78.5 & 82.6      & 69.7          & 57.1       & 59.5     & 69.5 \\
\bottomrule
\end{tabular}
}
\end{minipage}
\vspace{-5mm}
\end{table}

\paragraph{Recap.}
R-TAP consists of four key components: 
(a) GRPO-based reinforcement learning for Think-Answer trajectories, 
(b) a recursive reward mechanism derived from both the confidence generator and intermediate results, 
(c) suppression of unnecessary Oops-style refinement, and 
(d) improved inference efficiency resulting from reduced refinement steps. 

Importantly, while prior uncertainty- or refinement-based approaches use confidence signals for reranking, filtering, or verification after generation, R-TAP integrates confidence as an internal reinforcement signal during training. This signal continuously modulates recursive reasoning depth, directly shaping the learned reasoning policy instead of performing post-hoc correction at inference time.

\paragraph{Future Works.}
We plan to further advance R-TAP from the perspective of efficiency-oriented model design~\cite{lee2024collavo, lee2024moai, lee2024trol, lee2024phantom, lee2024vlsi, lee2025genrecal, lee2025unified, lee2025masking} and multiple evaluation benchmark~\cite{lee2025multiverse, lee2025refinebench}. In particular, we aim to develop adaptive recursion strategies that dynamically determine the necessity and depth of additional Think–Answer cycles, thereby minimizing redundant computation while preserving reasoning accuracy. Instead of relying on fixed recursion depth or static confidence thresholds, lightweight gating mechanisms~\cite{lee2022masking} or early-exit policies could be learned to selectively allocate reasoning steps based on estimated uncertainty. We also intend to explore parameter-efficient training schemes—such as partial fine-tuning or modular confidence heads—to reduce memory and training overhead, making R-TAP more practical for resource-constrained environments. Finally, extending confidence-guided recursive reasoning to smaller-scale models while maintaining competitive performance will be a key direction, enabling efficient yet reliable and robust~\cite{kim2023causal, kim2021distilling, kim2023demystifying, lee2023mitigating, lee2020towards} systems suitable for real-world deployment.


\end{document}